%% file: main.tex
\documentclass[letterpaper, 10 pt, conference]{ieeeconf}  

\IEEEoverridecommandlockouts                              

\usepackage{graphics} 
\usepackage{graphicx}
\usepackage{epsfig} 
\usepackage{mathptmx} 
\usepackage{times} 
\usepackage{amsmath} 

\usepackage{amssymb}  
\usepackage{bm}
\usepackage{algorithm}
\usepackage[noend]{algpseudocode}
\usepackage{caption}
\usepackage{subcaption}
\usepackage{physics}
\usepackage{xcolor}
\usepackage{tikz}
\usetikzlibrary{fit,
                positioning}
\usepackage[belowskip=1ex]{subcaption}  

\usetikzlibrary{calc} 
\usetikzlibrary{positioning}
\usepackage{balance}
\usepackage{soul}
\usetikzlibrary{arrows}
\usepackage{cite}
\usepackage{url}
\usepackage{multirow}
\usepackage{booktabs} 
\input{tex/preamble.tex}

\usepackage{setspace}
\title{\LARGE \bf
Real-time Contact State Estimation in Shape Control of Deformable Linear Objects under Small Environmental Constraints
}

\author{Kejia Chen$^{1}$, Zhenshan Bing$^{1}$, Yansong Wu$^{1}$, Fan Wu$^{1}$, Liding Zhang$^{1}$, Sami Haddadin$^{1}$,  Alois Knoll$^{1}$ 
\thanks{$^{1}$K. Chen, Z. Bing, Y. Wu, F. Wu, L. Zhang, S. Haddadin and A. Knoll are with the Department of Informatics, Technical University of Munich, Germany.
        {\tt\small kejia.chen@tum.de}}%
}

\begin{document}

\maketitle
\thispagestyle{empty}
\pagestyle{empty}

\begin{abstract}
Controlling the shape of deformable linear objects using robots and constraints provided by environmental fixtures has diverse industrial applications.
In order to establish robust contacts with these fixtures, accurate estimation of the contact state is essential for preventing and rectifying potential anomalies.
However, this task is challenging due to the small sizes of fixtures, the requirement for real-time performances, and the infinite degrees of freedom of the deformable linear objects.
In this paper, we propose a real-time approach for estimating both contact establishment and subsequent changes by leveraging the dependency between the applied and detected contact force on the deformable linear objects.
We seamlessly integrate this method into the robot control loop and achieve an adaptive shape control framework which avoids, detects and corrects anomalies automatically.
Real-world experiments validate the robustness and effectiveness of our contact estimation approach across various scenarios, significantly increasing the success rate of shape control processes.

\end{abstract}

\input{sections/sec1_intro.tex}

\input{sections/sec2_related.tex}
\input{sections/sec3_formulation.tex}

\input{sections/sec4_estimation.tex}

\input{sections/sec5_enhance.tex}

\input{sections/sec6_experiment.tex}
\input{sections/sec7_conclusion.tex}





\bibliographystyle{IEEEtran}
\bibliography{references}

\balance

\end{document}

%% file: tex/preamble.tex
\usepackage{amssymb}
\usepackage{amsfonts}
\usepackage{amsmath}
\usepackage{amsthm}
\usepackage{bm}

\input{tex/comments.tex}

\input{tex/symbols}

%% file: tex/comments.tex
\usepackage[normalem]{ulem}                                        
\usepackage{marginnote}
\usepackage[textwidth=10ex,colorinlistoftodos]{todonotes}

\colorlet{mh}{red}
\colorlet{fwu}{red}
\colorlet{ywu}{blue}
\colorlet{kchen}{blue}
\colorlet{lchen}{green}
\colorlet{zbing}{green}
\colorlet{shaddadin}{purple}
\colorlet{iperez}{cyan}
\colorlet{schneider}{magenta}



%% file: tex/symbols.tex
\usepackage{amssymb}
\usepackage{mathtools}
\usepackage{sansmath}

\mathchardef\mhyphen="2D   

\newcommand{\RNum}[1]{\uppercase\expandafter{\romannumeral #1\relax}}

%% file: sections/sec1_intro.tex
\section{INTRODUCTION}

Controlling the shape of deformable linear objects (DLOs) with robot manipulators has a wide range of industrial applications, such as cable routing \cite{chen2023contactaware}, wire-harness assembly in manufacturing \cite{navas2022wire}, or manipulation of endoscopes in robotic surgeries~\cite{haouchine2018vision}.
These shape control tasks pose a significant challenge due to the inherent mismatch between the finite constraints that can be imposed on DLOs by manipulators and the infinite degrees of freedom DLOs possess~\cite{huang2023learning}.
In tackling this challenge and striving to achieve complex shapes, additional constraints are required, typically provided by contacts from environmental fixtures~\cite{zhu2019robotic,huo2022keypoint, jin2022robotic, waltersson2022planning, suberkrub2022feel, chen2023contactaware}.
To safely fasten DLOs using fixtures, reliable contact state estimation is crucial for prevention of potential anomalies such as misalignment or insufficient pushing, and for enhancing the overall robustness of the shape control system.

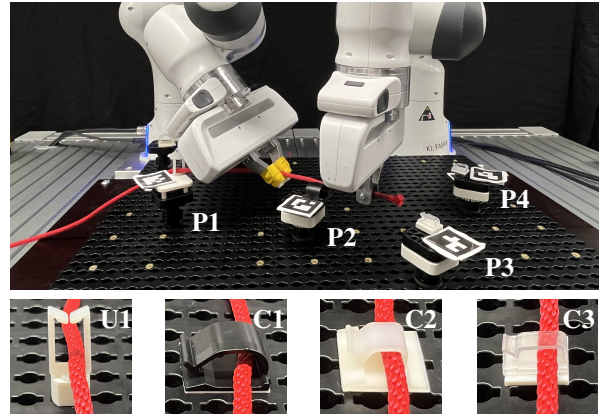
\begin{figure}[t]
    \centering
    \begin{tikzpicture}
    

    \node[inner sep=0pt] (russell) at (-2.5,0)
    {\includegraphics[width=0.44\textwidth]{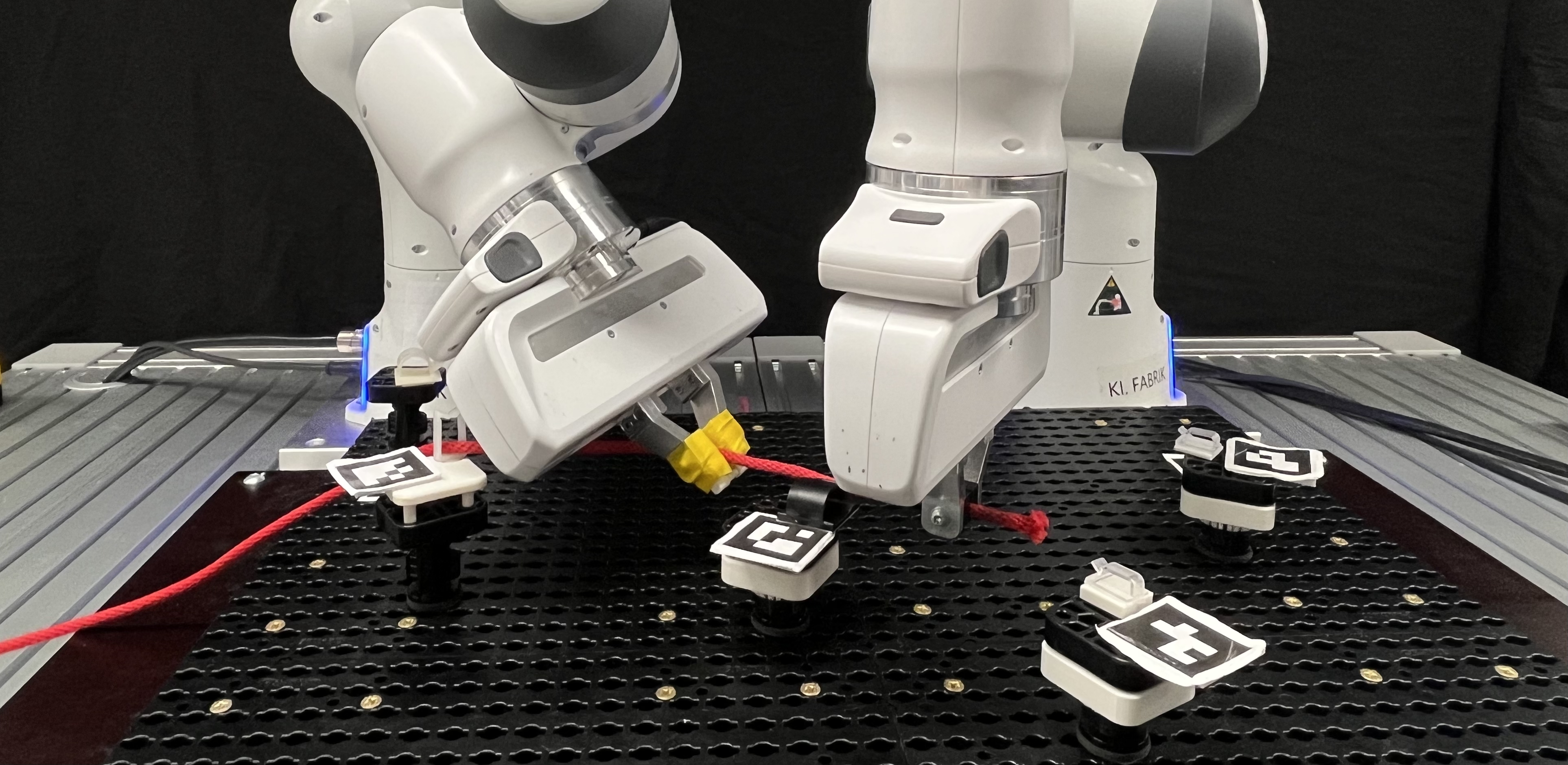}};
    \node [text=white] at (-3.8, -1) {\textbf{P1}};
    \node [text=white] at (-2, -1.2) {\textbf{P2}};
    \node [text=white] at (0.1, -1.6) {\textbf{P3}};
    \node [text=white] at (0.3, -0.7) {\textbf{P4}};
    \node[inner sep=0pt] (russell) at (-5.6,-2.8)
    {\includegraphics[width=0.09\textwidth]{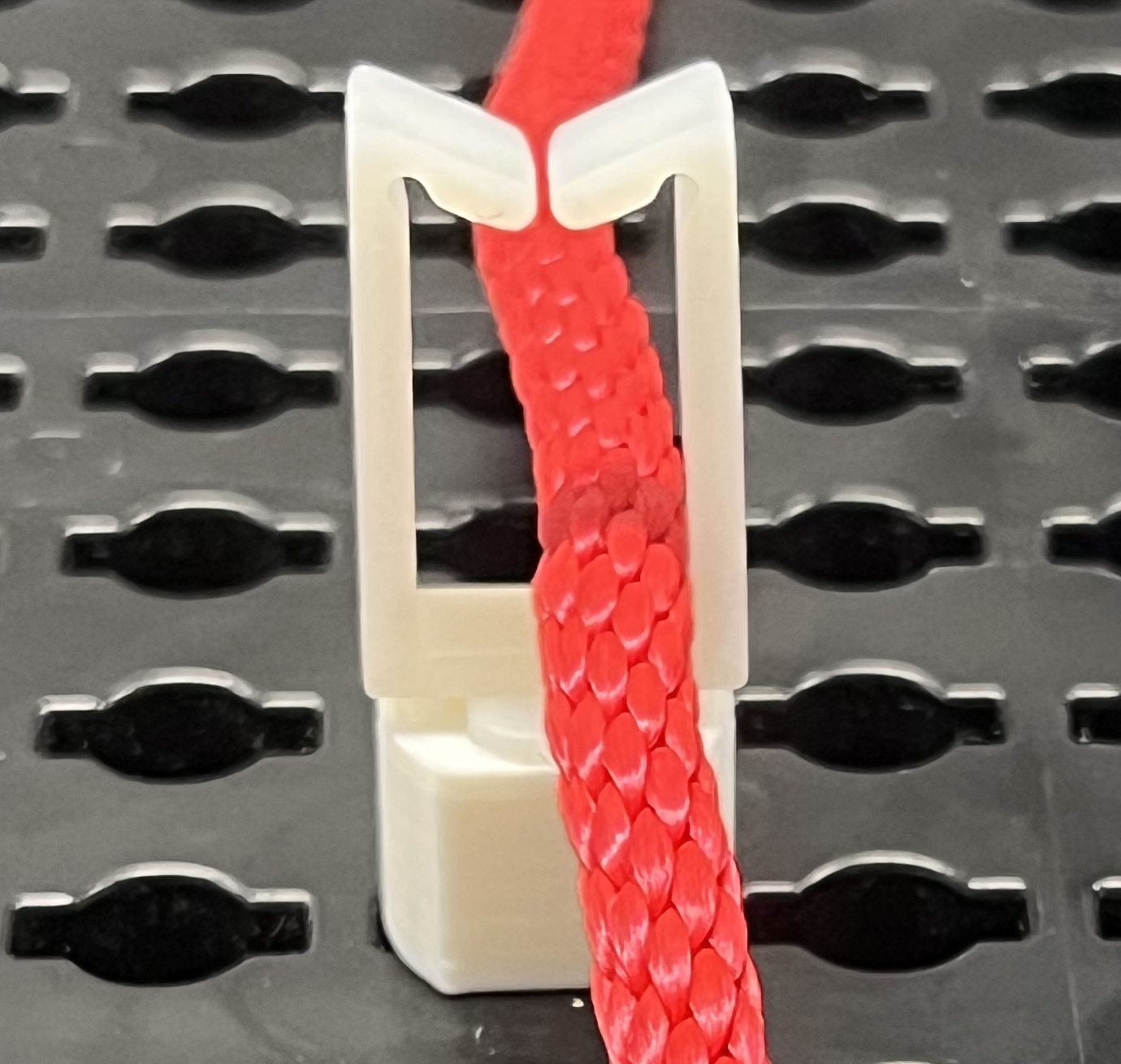}};
    \node [text=white] at (-5.0,-2.3) {\textbf{U1}};
    \node[inner sep=0pt] (russell) at (-3.54,-2.8)
    {\includegraphics[width=0.09\textwidth]{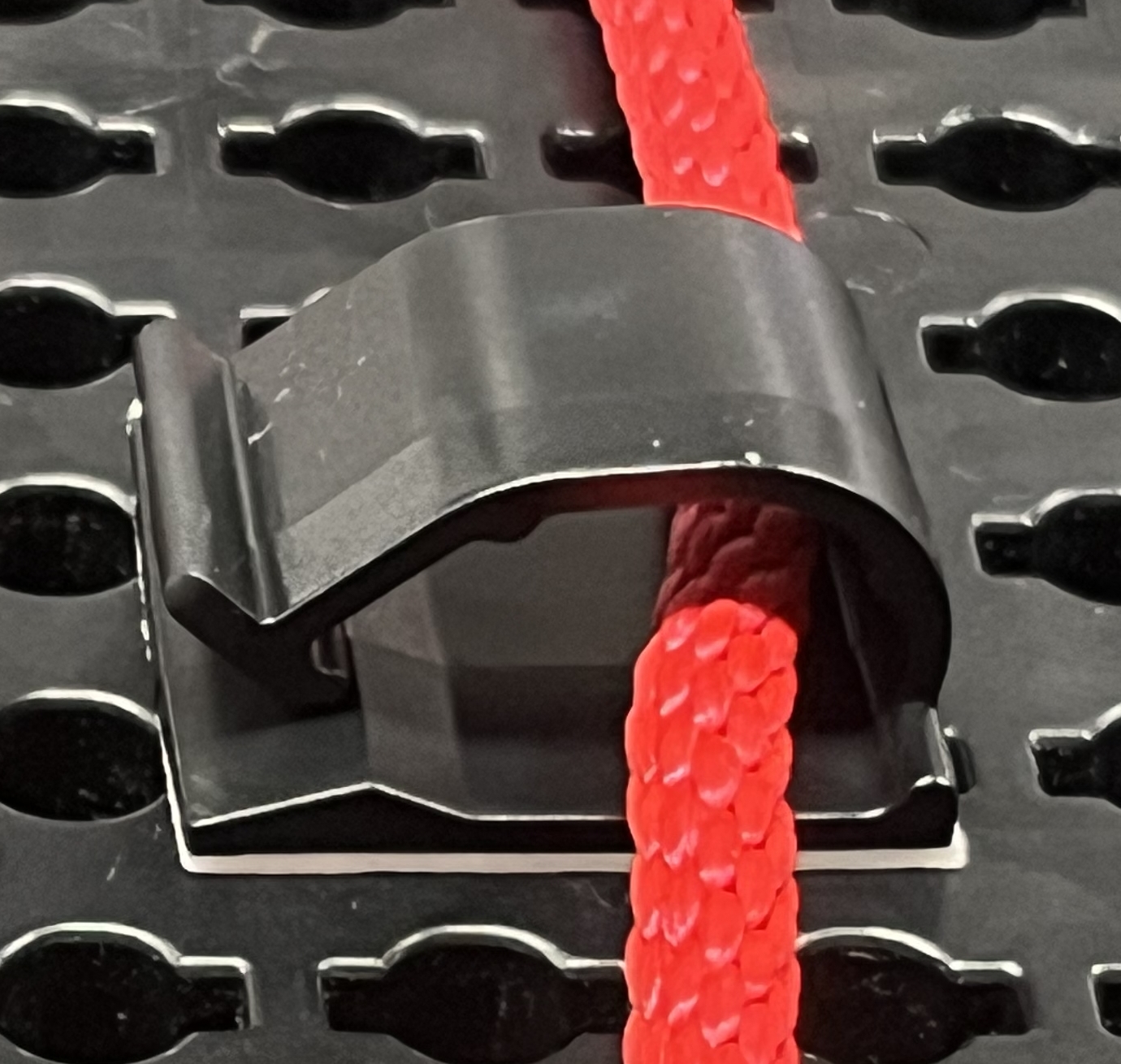}};
    \node [text=white] at (-2.97,-2.3) {\textbf{C1}};
    \node[inner sep=0pt] (russell) at (-1.48,-2.8)
    {\includegraphics[width=0.09\textwidth]{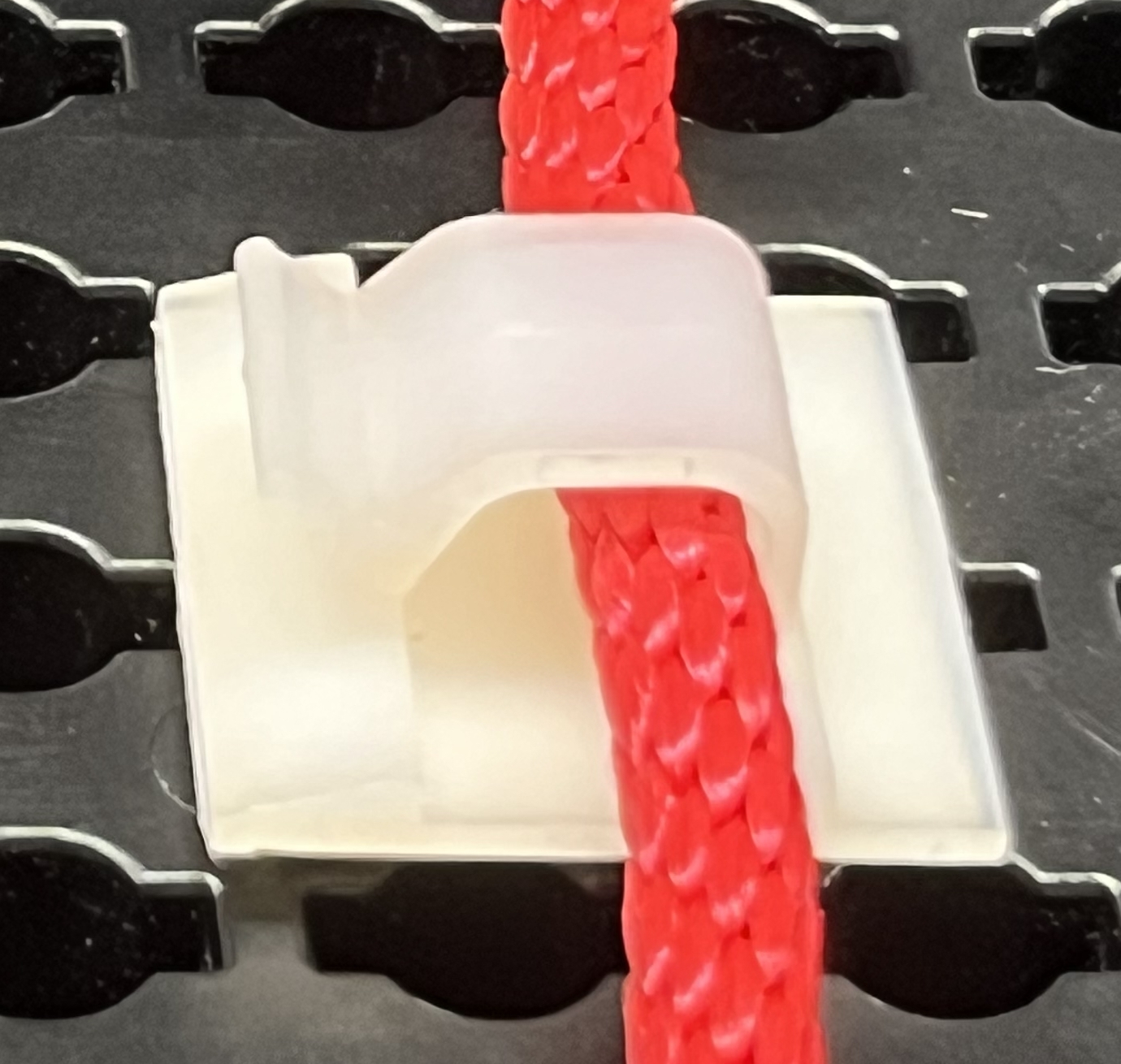}};
    \node [text=white] at (-0.95,-2.3) {\textbf{C2}};
    \node[inner sep=0pt] (russell) at (0.6,-2.8)
    {\includegraphics[width=0.09\textwidth]{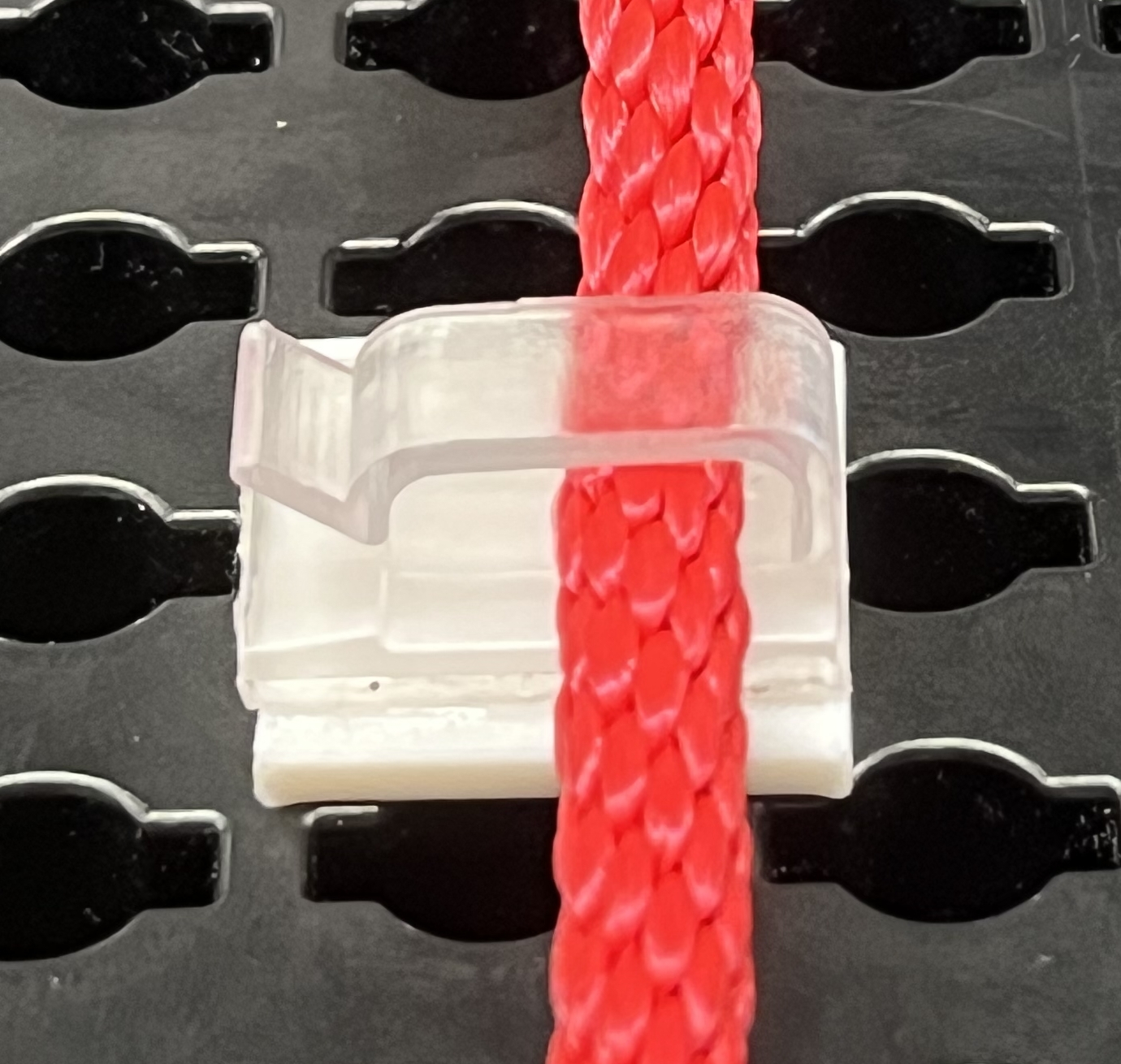}};
    \node [text=white] at (1.15,-2.3) {\textbf{C3}};
    
    \end{tikzpicture}
    \caption{Setup. Top: shape control of DLO with 4 environmental fixtures using two robots. Bottom: various clip fixtures. }
    \label{fig: clips}
    \vspace{-1.7em} 
\end{figure}

\begin{figure}[t]
    \centering
    \begin{tikzpicture}
    \node[inner sep=0pt] (russell) at (-3.5,1.87)
    {\includegraphics[width=0.225\textwidth]{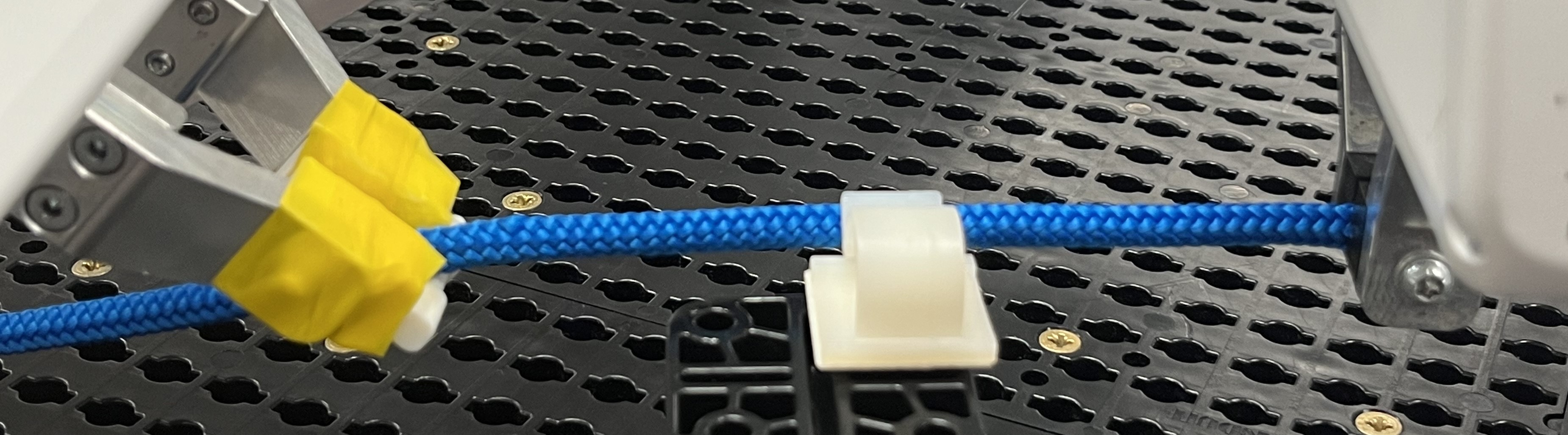}};
    \node[inner sep=0pt] (russell) at (-3.5,0.63)
    {\includegraphics[width=0.225\textwidth]{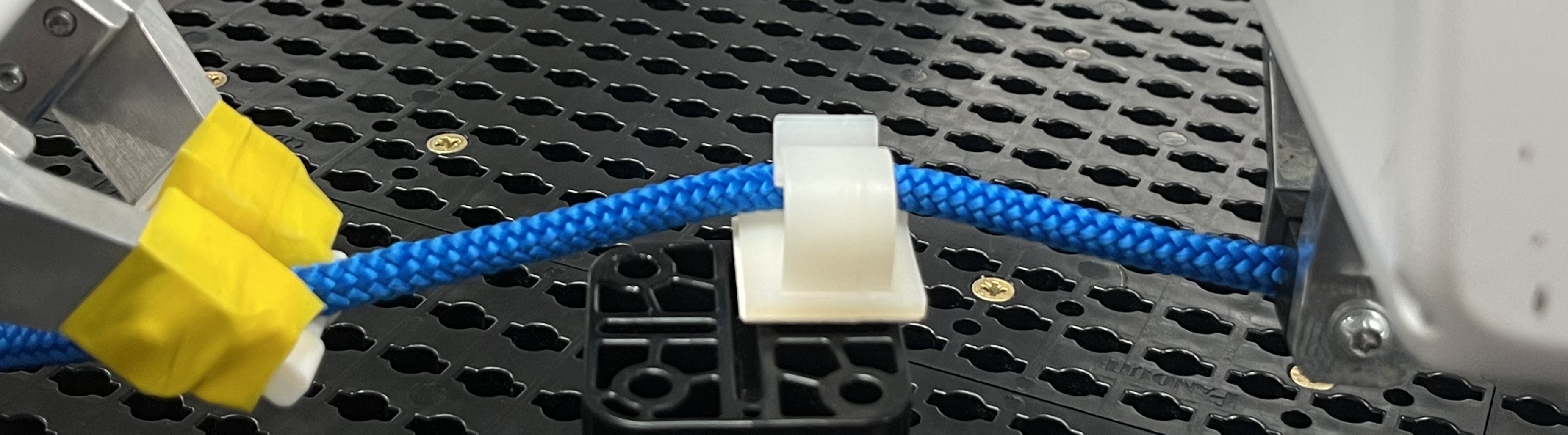}};
    \node[inner sep=0pt] (russell) at (0.7,1.25)
    {\includegraphics[width=0.22\textwidth]{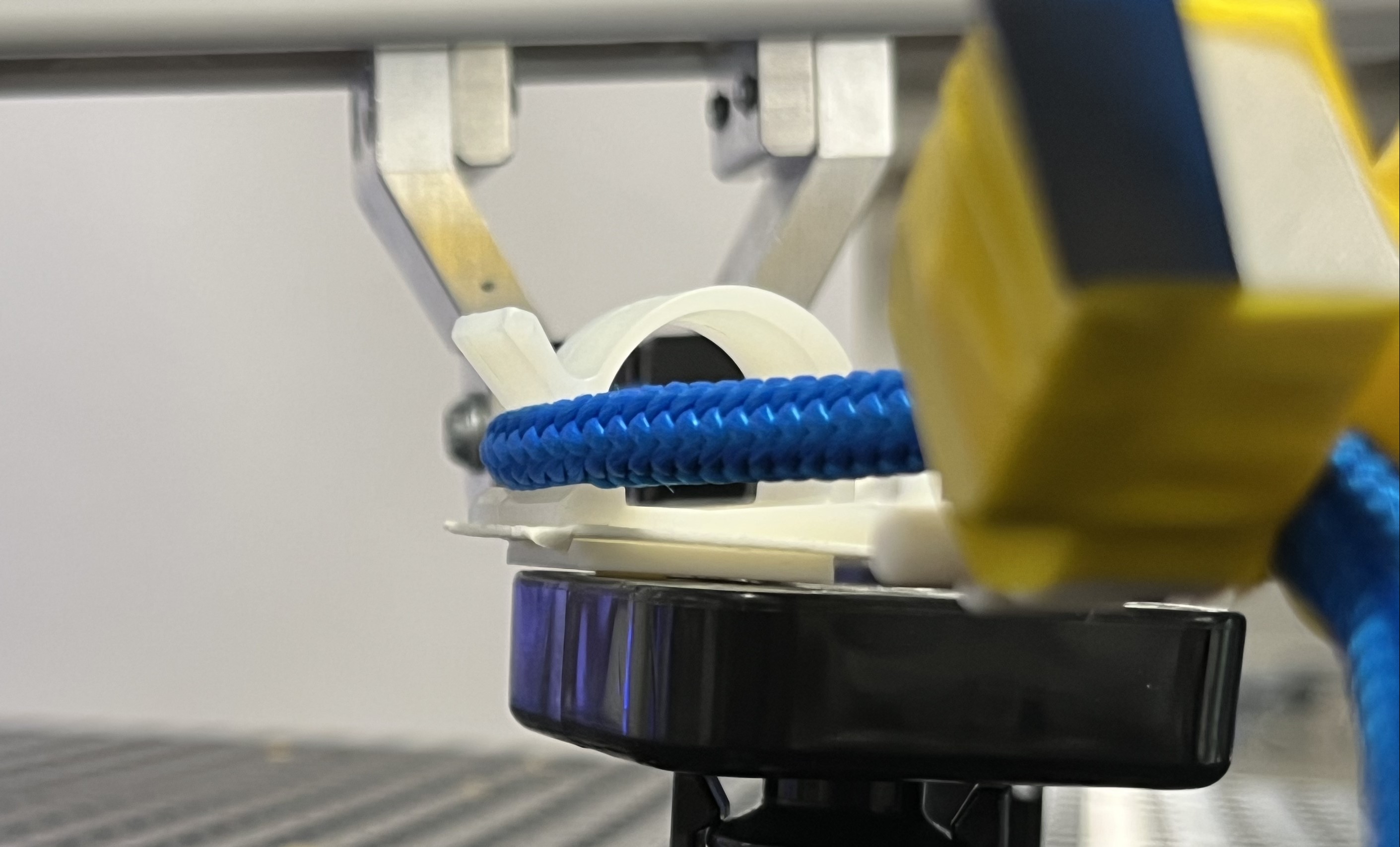}};
    \node at (-3.5,-2.71) {(a)};

    \node[inner sep=0pt] (russell) at (-3.5,-0.63)
    {\includegraphics[width=0.225\textwidth]{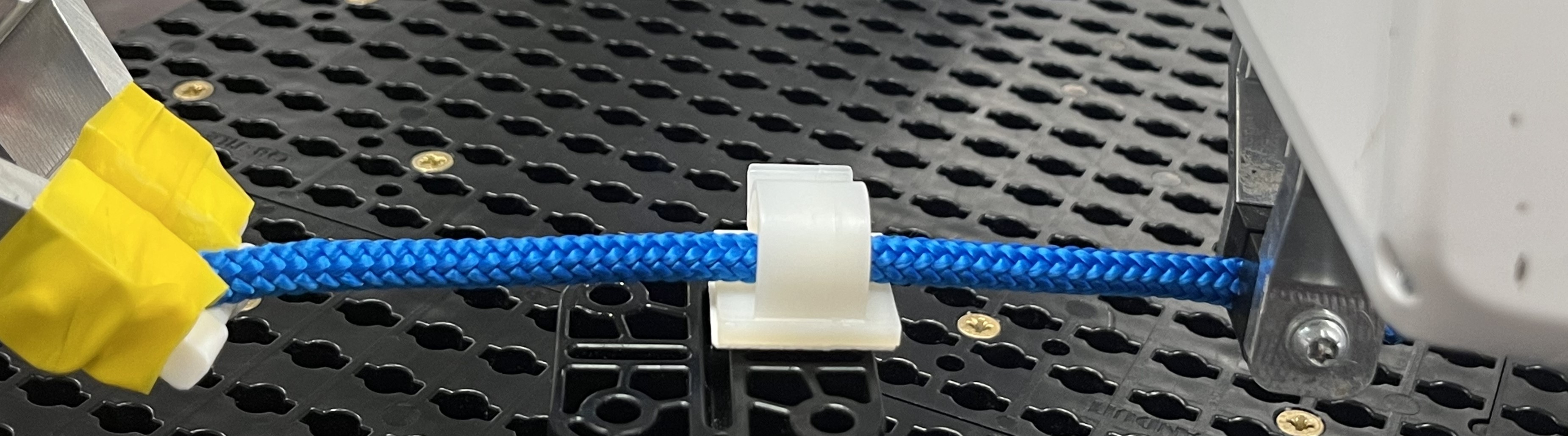}};
    \node[inner sep=0pt] (russell) at (-3.5,-1.90)
    {\includegraphics[width=0.225\textwidth]{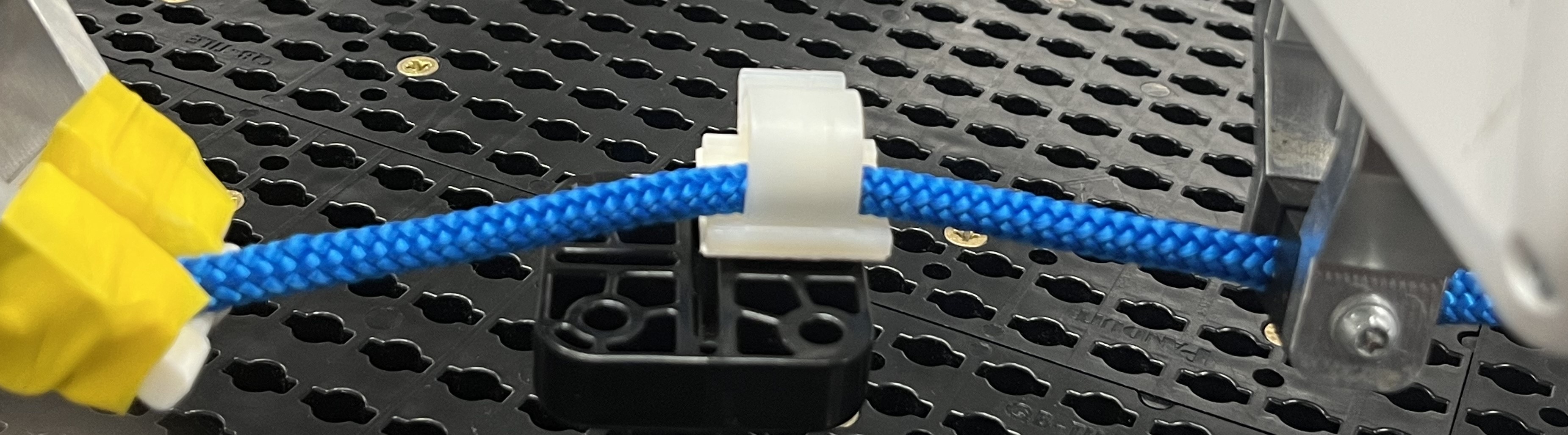}};
    \node[inner sep=0pt] (russell) at (0.7,-1.26)
    {\includegraphics[width=0.22\textwidth]{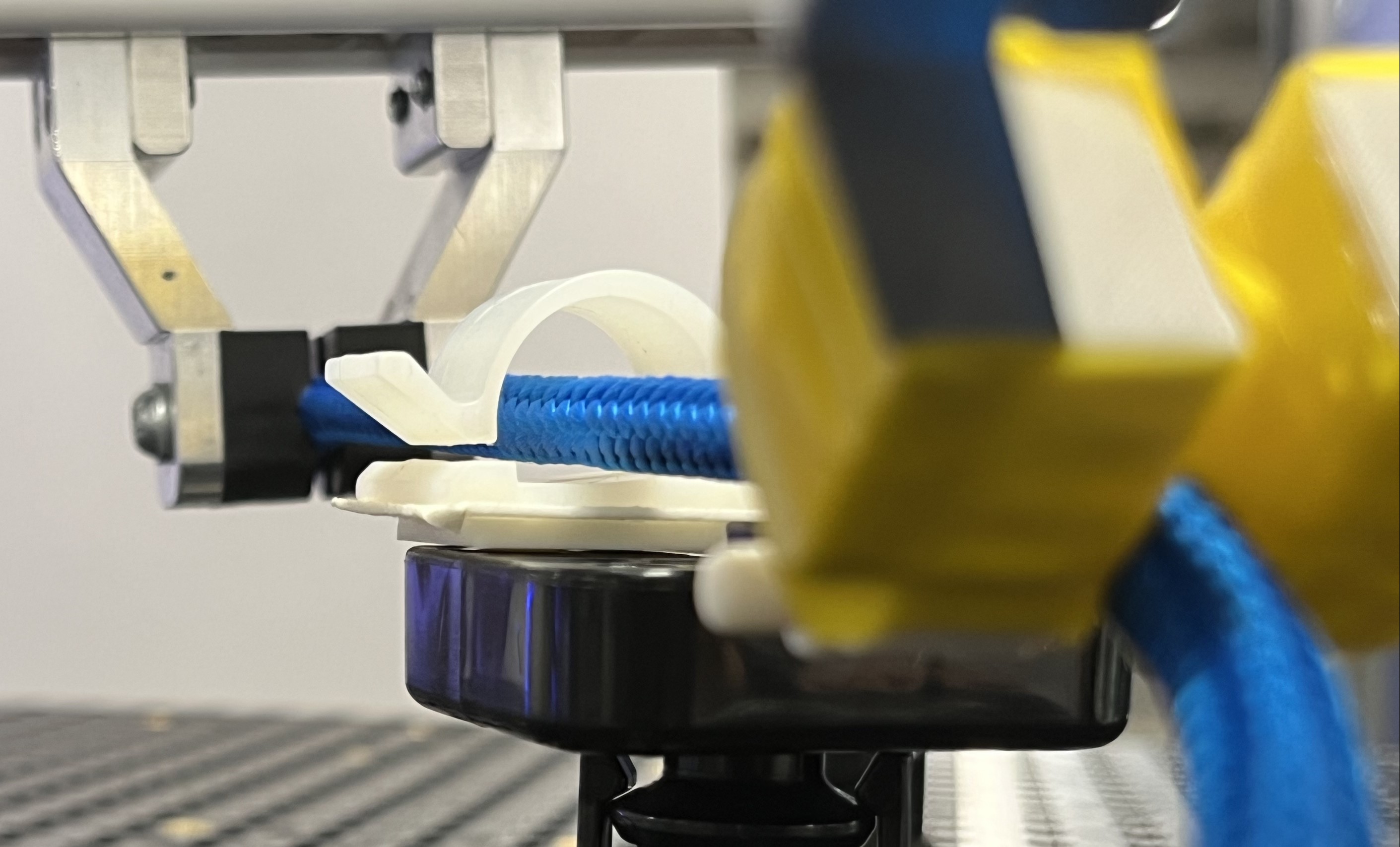}};
    \node at (0.7,-2.71) {(b)};
    
    \end{tikzpicture}
    \caption{Clip fixing and DLO deformation. (a) Top views. From top to bottom: contact-insertion-fixed-overforce movement. (b) Left views. Top: insertion; bottom: fixed.}
    \label{fig: clip_fixing_example}
    \vspace{-1.7em} 
\end{figure}

Taking typical fixtures which are widely used in wire-harness assembly as an example, different fixtures are designed to provide different types of contact for the DLO.
For pillar-like fixtures~\cite{zhu2019robotic} or channel-like fixtures~\cite{jin2022robotic}, the contact states are simply binary, only indicating whether contact has been established or not.
In contrast, due to their own deformations, clip-like fixtures depicted in Fig~\ref{fig: clips} introduce a dynamic and more complex contact process~(see Fig~\ref{fig: clip_fixing_example}):
as the DLO advances towards the clip, it initially makes contact with the clip's opening.
Subsequently, as the DLO is pushed inward, the clip is forced to open to let the object in. 
Once the object is securely fastened inside the clip, the contact is detached, unless the object moves further and collides with the rear part of the clip.
Despite the brevity of this process in terms of both time and displacement, the contact force undergoes multiple distinct stages of change, notably featuring an instantaneous and abrupt drop at the moment of insertion.
Similar contact patterns can also be observed during suturing or tissue retraction of minimally invasive surgery~\cite{jansen2009surgical}, when a surgical tool pierces from one type of tissue to another.


While contact estimation seems to be an intuitive task for humans due to their superior tactile sense, it is challenging when performed by robots.
Existing contact estimation approaches typically rely on visual perception or robot motion information.
However, they are not reliable to be applied to contacts in clip fixing scenarios.
Firstly, the small size of fixtures and the resulting limited displacements of robots during clip fixing hinders the effectiveness of vision-based methods.
On one hand, it demands highly precise object segmentation and tracking.
On the other hand, it imposes strict real-time requirements for contact estimation, requiring computations to match the robot control loop frequency.
As a result, visual perception algorithms used in prior works~\cite{zhu2019robotic, jin2022robotic, huo2022keypoint} are not practical due to their dependence on slower image processing. 
Secondly, the contact is established not with the robot but with the DLO itself,  which makes direct contact measurement impossible~\cite{zhu2019robotic}.
In cases involving rigid objects or direct contact between robots and fixtures, contact interactions can be characterized easily by robot displacement~\cite{yao2023estimating}:
the displacement of the robot will pause temporarily upon blockages, and only resumes once the object is inserted in.
However, in the case of a DLO, its contact state lags behind the robot motion due to the deformation.
For example, in Fig~\ref{fig: clip_fixing_example}(a), despite that we have stretched the grasped DLO to be tension, noticeable deformation still exists, and the two robots may continue advancing even when DLO is still blocked. 
In short, vision-based and motion-based approaches are rendered impractical, which motivates us to seek a more efficient and robust approach for contact estimation.

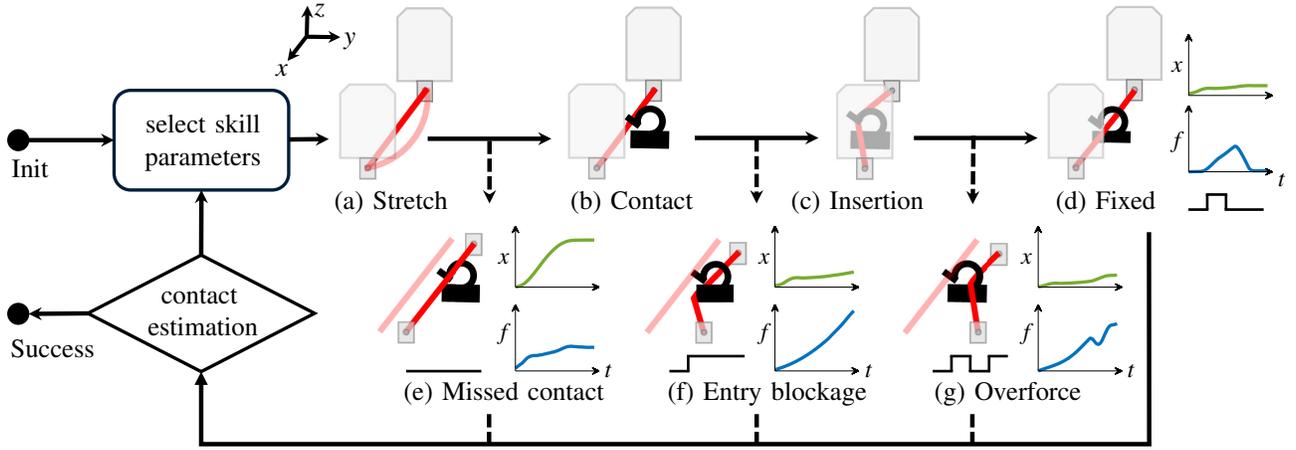
\begin{figure*}[t!]
     \centering
    \begin{tikzpicture}
    \node[inner sep=0pt] (russell) at (-0.1,0)
    {\includegraphics[width=0.94\textwidth]{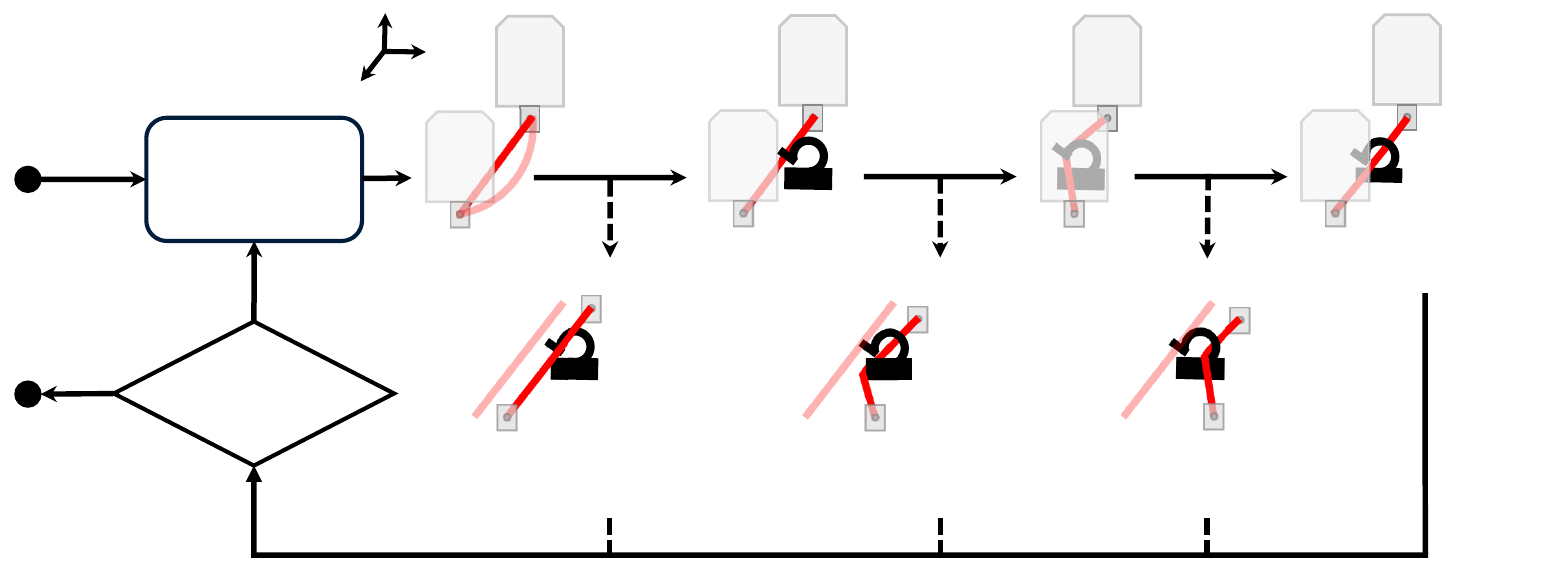}};
    \node at (-4.65, 2.1) {$x$};
    \node at (-3.75, 2.5) {$y$};
    \node at (-4.15, 2.9) {$z$};
    
    \node[inner sep=0pt] (russell) at (7.9,1.15)
    {\includegraphics[width=0.08\textwidth]
    {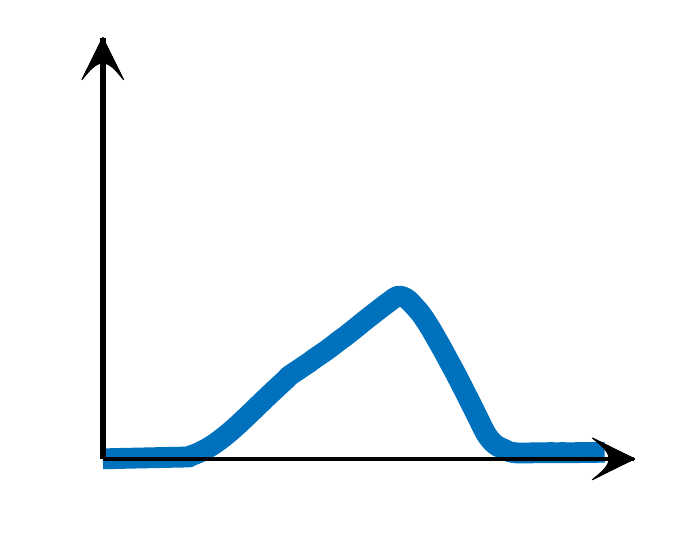}};
     \node[inner sep=0pt] (russell) at (7.9,2.2)
    {\includegraphics[width=0.08\textwidth]
    {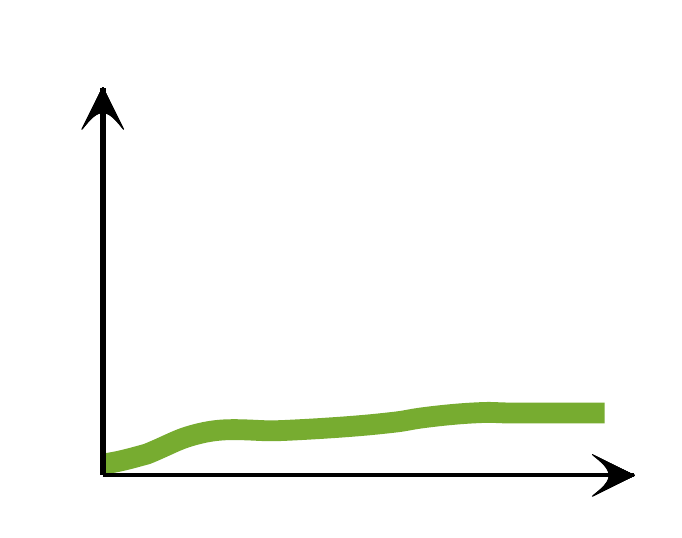}};
    \node at (8.65,0.65) {\small{$t$}};
    \node at (7.25,1.15) {\small{$f$}};
    \node at (7.25,2.15) {\small{$x$}};
    \draw [-][line width=1pt] (7.4,0.25)--(7.667,0.25);
    \draw [-][line width=1pt] (7.65,0.25)--(7.65,0.45);
    \draw [-][line width=1pt] (7.632,0.45)--(7.917,0.45);
    \draw [-][line width=1pt] (7.9,0.25)--(7.9,0.45);
    \draw [-][line width=1pt] (7.882,0.25)--(8.4,0.25);
    
    \node[inner sep=0pt] (russell) at (-1.05,-1.5)
    {\includegraphics[width=0.08\textwidth]
    {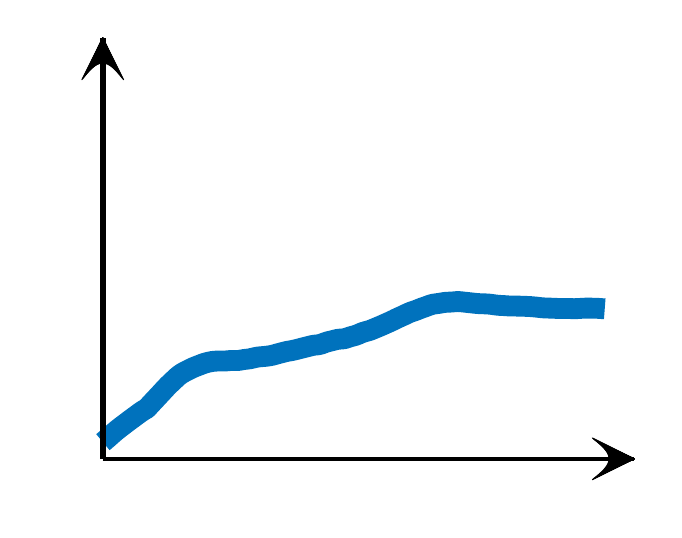}};
     \node[inner sep=0pt] (russell) at (-1.05,-0.35)
    {\includegraphics[width=0.08\textwidth]
    {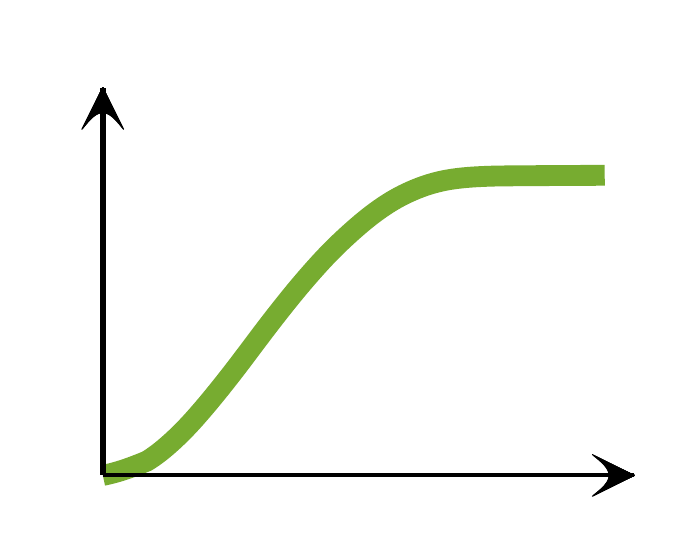}};
    \node at (-0.3,-1.9) {\small{$t$}};
    \node at (-1.7,-1.4) {\small{$f$}};
    \node at (-1.7,-0.4) {\small{$x$}};
    \draw [-][line width=1pt] (-3,-1.9)--(-2,-1.9);

    \node[inner sep=0pt] (russell) at (2.4,-1.5)
    {\includegraphics[width=0.08\textwidth]
    {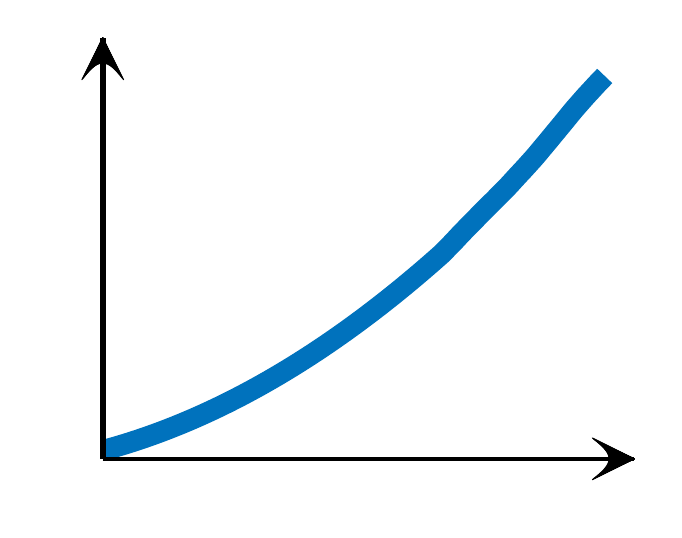}};
     \node[inner sep=0pt] (russell) at (2.4,-0.35)
    {\includegraphics[width=0.08\textwidth]
    {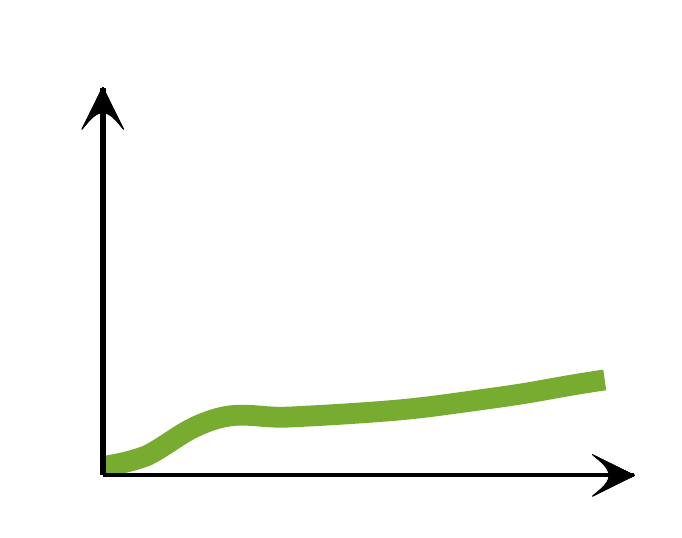}};
    \node at (3.16,-1.9) {\small{$t$}};
    \node at (1.76,-1.4) {\small{$f$}};
    \node at (1.76,-0.4) {\small{$x$}};
    \draw [-][line width=1pt] (0.5,-1.9)--(0.767,-1.9);
    \draw [-][line width=1pt] (0.75,-1.9)--(0.75,-1.7);
    \draw [-][line width=1pt] (0.732,-1.7)--(1.5,-1.7);
    
    \node[inner sep=0pt] (russell) at (5.9,-1.5)
    {\includegraphics[width=0.08\textwidth]
    {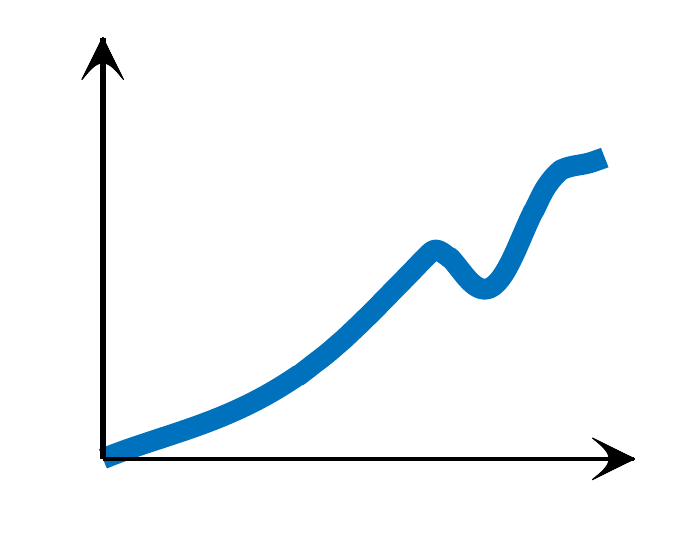}};
     \node[inner sep=0pt] (russell) at (5.9,-0.35)
    {\includegraphics[width=0.08\textwidth]
    {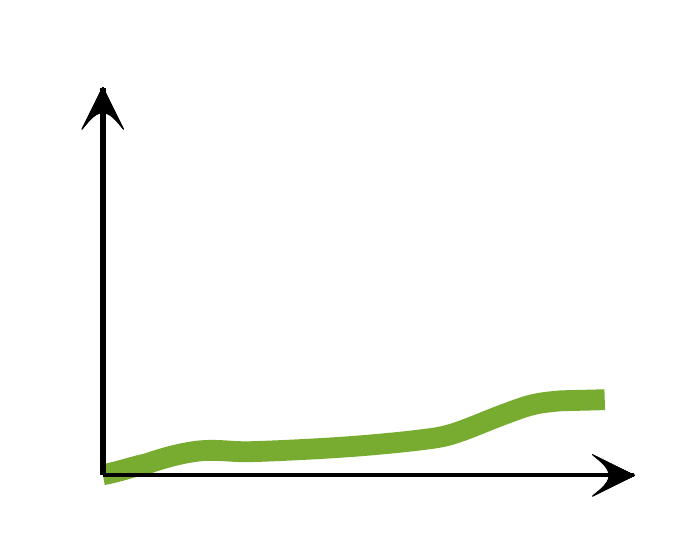}};
    \node at (6.65,-1.9) {\small{$t$}};
    \node at (5.25,-1.4) {\small{$f$}};
    \node at (5.25,-0.4) {\small{$x$}};
    \draw [-][line width=1pt] (4,-1.9)--(4.268,-1.9);
    \draw [-][line width=1pt] (4.25,-1.9)--(4.25,-1.7);
    \draw [-][line width=1pt] (4.232,-1.7)--(4.518,-1.7);
    \draw [-][line width=1pt] (4.5,-1.9)--(4.5,-1.7);
    \draw [-][line width=1pt] (4.482,-1.9)--(4.767,-1.9);
    \draw [-][line width=1pt] (4.75,-1.9)--(4.75,-1.7);
    \draw [-][line width=1pt] (4.732,-1.7)--(5,-1.7);

    \node at (-5.7,1.4) {select skill};
    \node at (-5.7,0.9) {parameters};
    
    \node at (-5.75,-0.9) {contact};
    \node at (-5.7,-1.3) {estimation};

    \node at (-8.0,0.8) {Init};
    \node at (-7.7,-1.6) {Success};

    \node at (-3.2, 0.35) {(a) Stretch};
    \node at (0.0, 0.35) {(b) Contact};
    \node at (3.0,0.35) {(c) Insertion};
    \node at (6.3,0.35) {(d) Fixed};
    \node at (-1.7,-2.2) {(e) Missed contact};
    \node at (1.8,-2.2) {(f) Entry blockage};
    \node at (5.0,-2.2) {(g) Overforce};
    \end{tikzpicture}
    \vspace{-0.5em} 
    \caption{Clip fixing process. The red curve represents the DLO. The black fixture represents the clip. The gray polygon represents the robot hand and finger tips. (a), (b), (c) and (d) in the first row describe the ideal clip fixing process. (e), (f), (g) in the second row describe the failures which may happen at different stages. The tendency of displacement (green curve) and contact force (blue curve) are depicted next to each failure. For simplicity, the hand is omitted in the second row.}
    \label{fig: overview}
    \vspace{-1.5em} 
\end{figure*}

Inspired by the pivotal role of tactile information in human perception of deformable objects, and considering its effortless acquisition by robot sensors within each control loop, we present a real-time method based on contact forces that can accurately estimate contact state of DLOs subject to various environmental constraints. 
Through an in-depth analysis of contact characteristics exhibited by fixtures as well as their interaction with DLO, we design two indicators for robust contact state estimation. 
Subsequently, we integrate these indicators into the clip fixing skill developed in our prior work~\cite{chen2023contactaware}, and realize a self-adjusting DLO shape control framework.
This framework shows the capability to dynamically adapt to varying contact scenarios.
The contributions of this work are summarized as follows.
\begin{itemize}
    \item Based on the dependency between the applied and the detected contact force, we propose a contact establishment indicator and a contact change indicator. These two indicators describe the initial establishment and the following change of DLO's contact against fixtures robustly across various settings.
    \item Through these indicators, we develop a contact state estimation method which could run in real time in $1$ kHz control loop and accurately detect potential anomalies in the shape control process.
    \item We integrate the contact state estimation method into the DLO shape control framework, which dynamically adjusts its parameters in case of anomalies. Real-world experiments have validated the improvements in performance of DLO shape control process with fixtures after this integration.
\end{itemize}

%% file: sections/sec2_related.tex
\section{Related Work}

The utilization of environmental fixtures to control the shape of DLOs was first introduced by Zhu et al.~\cite{zhu2019robotic}. 
Since then, there has been a widespread adoption of this approach in DLO manipulation~\cite{huo2022keypoint, jin2022robotic, waltersson2022planning, suberkrub2022feel, huang2023learning, chen2023contactaware}.
A crucial step in this process is estimating the contact status of the DLO constrained by fixtures. 

Existing research in contact state estimation for deformable objects often focuses on constructing a contact distribution, encompassing both contact location and the magnitude of contact forces. 
This distribution could be built and updated online during the manipulation, for example, by biomechanical mapping based on 3D reconstruction from stereo endoscopic images of an organ~\cite{haouchine2018vision}, or volumetric stiffness field modeling when touching artificial plants~\cite{yao2023estimating}.
Alternatively, it can be generated by neural networks trained on a combination of visual, tactile, and robot movement data beforehand~\cite{erickson2017does, wang2022visual, wi2022virdo}.

However, when dealing with DLO manipulation using fixtures, constructing a complete contact distribution becomes impractical for several reasons. 
Unlike direct tactile sensing on deformable objects, in DLO manipulation, the contact force acts on the DLO itself, making precise measurement challenging~\cite{zhu2019robotic}. 
Furthermore, traditional distance threshold-based contact detection methods used in prior research~\cite{haouchine2018vision, yao2023estimating} are less robust due to the dynamic deformations exhibited by the DLO during manipulation.
Finally, DLO manipulation primarily involves contact with only a small segment of the DLO, which makes focusing on the existence and position of contact points more suitable than creating a contact distribution.
Therefore, previous works in manipulation of DLOs tend to adopt more intuitive and computationally efficient criteria to estimate the contact states, either by detecting the contact establishment and resulting shape changes from top images~\cite{zhu2019robotic, huo2022keypoint, jin2022robotic, waltersson2022planning, huang2023learning}, or by detecting the contact position on the DLO from force measurements~\cite{ suberkrub2022feel, chen2023contactaware}. 

Following this idea, our paper introduces two straightforward yet effective indicators for estimating DLO's contact state during their manipulation with environmental fixtures. 
These indicators not only detect contact establishment but also identify contact changes afterwards in multi-stage contact processes.
Importantly, our method demonstrates superior real-time performance and can be seamlessly integrated into the robot control loop and DLO shape control process.

%% file: sections/sec3_formulation.tex
\section{Problem Formulation}
\subsection{Clip Fixing Process}~\label{subsec: process}
We formulate the clip fixing process based on the clip-fixing skill introduced in our prior work~\cite{chen2023contactaware}.
As is shown in Fig~\ref{fig: overview}(a), the movement of robots as well as the cable is described in an object-centered coordinate frame: 
the $x$-axis corresponds to the tangential direction of the cable, 
the $y$-axis represents the insertion direction of the clip, 
and the $z$-axis indicates the opening (actuation) direction of the clip.

As illustrated in Fig~\ref{fig: overview}, the clip fixing process is defined as a directed transition graph of manipulation primitives (MPs).
Each MP consists of a desired linear velocity $\mathbf{\dot{x}^{d}} \in \mathbb{R}^3$ and feedforward force $\mathbf{f} \in \mathbb{R}^3$, both controlled under an adaptive impedance controller~\cite{johannsmeier2019framework}.
Initially, both robots securely grasp each end of a DLO segment. 
As two robot exert forces $\mathbf{f_{stretch}}=[\pm f_{stretch}, 0, 0]^T$ in the opposite direction, the segment is stretched until it becomes tension (Fig.\ref{fig: overview}(a)). 
This stretching force is maintained throughout the subsequent stages. 
Following this, robots guide the segment to move along $y$-axis ($\mathbf{\dot{x}}^{d}=[0, 1, 0]^T$) to establish contact with the clip (Fig.\ref{fig: overview}(b)). 
Upon contact detection, robots insert the segment into the clip by applying a pushing force $\mathbf{f_{push}}=[0, f_{push}, 0]^T$ (Fig.\ref{fig: overview}(c)).
Once the DLO segment is fully inserted, the robots cease applying forces and further motion (Fig.\ref{fig: overview}(d)).


Although experiments in \cite{chen2023contactaware} have substantiated the effectiveness and advantages of the clip-fixing skill above for controlling and maintaining the DLO's shape, there exists some issues which may diminish the framework's robustness:
\begin{itemize}
    \item \textbf{Missed contact} (Fig~\ref{fig: overview}(e)). If the grasped DLO passes over the clip opening, no contact is established between the object and the clip. 
    Consequently, the DLO continues to move forward alongside the clip. 
    \item \textbf{Entry blockage} (Fig~\ref{fig: overview}(f)). If the grasped DLO moves below the clip opening, it will be blocked by the fixture base as the robots move further. 
    Once contact is established, it is maintained until the skill exits. 
    Similar phenomenons can be observed if $f_{push}$ is inadequate to overcome the elastic force $f_d$ of the clip.
    \item \textbf{Overforce movement} (Fig~\ref{fig: overview}(g)).  Excessive applied force $f_{push}$ or a delay in force removal following insertion can cause the DLO to continue moving forward, eventually colliding with the clip's rear end. 
    In extreme cases, this may result in damage.
\end{itemize}
We notice that these anomalies happen at different stages of the clip fixing process with different contact patterns, and can be detected and avoided by accurate estimation of DLO's current contact state with the clip.
\subsection{Contact with Clip}\label{subsec: clip_dynamics}

In contrast to static pillar or channel fixtures, the clip fixture exhibits elastic deformation when subjected to a pushing force at its opening. Consequently, the contact force exerted by the clip fixture is expected to be dynamic, multi-stage, and anisotropic. To gain a deeper insight into the contact patterns, we study force interactions with the clip.

\begin{figure}[t]
    \centering
    \begin{tikzpicture}
    \node[inner sep=0pt] (russell) at (-1.4,0)
    {\includegraphics[width=0.28\textwidth]{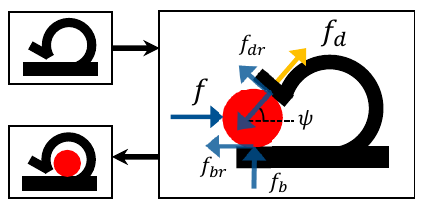}};
    \node at (-3.15,0) {(a)};
    \node at (-3.15,-1.35) {(c)};
    \node at (-0.6,-1.35) {(b)};
    \node[inner sep=0pt] (russell) at (2.7,0.05)
    {\includegraphics[width=0.17\textwidth]
    {figures/overview/fit_f_ext_proj.pdf}};
    \node at (2.9,-1.35) {(d)};
    \node at (2.9,-1) {\small{$x$ (m)}};
    \node[rotate=90] at (1.45,0) {\small{$f_d$ (N)}};
    \end{tikzpicture}
    \vspace{-0.5em} 
    \caption{Clip dynamics (a) before, (b) during, and (c) after insertion. (d) The change of contact force in this process.}
    \label{fig: clip_dynamics}
    \vspace{-1.7em} 
\end{figure}

The initial state of the clip is shown in Fig~\ref{fig: clip_dynamics}(a).
When a force $\mathbf{f}$ is applied to push an object into the clip, the clip performs deformation $\Delta \mathbf{h(t)}$ and thus applies an elastic force $\mathbf{f_{d}}(t)$ on the object (Fig~\ref{fig: clip_dynamics}(b)).
For simplicity, we assume that the clip opening possesses a constant stiffness denoted as $ K_{clip}$ and there exists a quasi-linear mapping $\mathbf{P}$ from the object's displacement $\mathbf{x}(t)$ to deformation $\mathbf{h}(t)$.
In this case, the elastic force can be formulated as
\begin{equation}
    \mathbf{f_{d}}(t)= K_{clip} \cdot \Delta \mathbf{h}(t) =  K_{clip} \cdot \Delta \mathbf{P}(\mathbf{x}(t))\textbf{.}
\end{equation}
When the magnitude of $\mathbf{f}$ is sufficiently large, the object is inserted into the clip and the contact is detached, causing both $\mathbf{h}(t)$ and $\mathbf{f_{d}}$ to become zero (Fig~\ref{fig: clip_dynamics}(c)).
The change of the magnitude of elastic force $f_{d}$ to displacement in this process is visualized in Fig~\ref{fig: clip_dynamics}(d).
In the pushing process, the elastic force initially increases, then experiences an abrupt drop to zero, and finally remains there until the skill terminates.

We then study the contact force applied on the object by the clip. 
Apart from $\mathbf{f_d}$, the object also experiences a support force from the fixture bottom $\mathbf{f_{b}}$ as well as two friction forces $\mathbf{f_{br}}$ and $\mathbf{f_{dr}}$. 
The total contact force applied on the object by the clip is thus a sum $\mathbf{f_c} =  \mathbf{f_{d}}+ \mathbf{f_{dr}} + \mathbf{f_{b}}+ \mathbf{f_{br}}$.
Consider only the components in $y$ direction, we have
\begin{equation} \label{equ: clip contact}
    f^y_{c} = {f_{d}} \cdot (2\mu\cdot \cos{\psi} + (1-\mu^2)\cdot \sin{\psi})\textbf{,}
\end{equation}
where $f_{d}$ represents the magnitude of $\mathbf{f_{d}}$.
Note that in the following sections, we adopt the same convention that bold variables (e.g., $\mathbf{v}$) represent vectors, while their scalar counterparts (e.g., $v$) denote the magnitude of these vectors.
$\psi$ is the angle and $\mu$ is the coefficient of friction.
In the clip fixing process, the change in $\psi$ is small and can be neglected. 
Thus, we can conclude that the contact force applied on the object $f^y_{c}$ is approximately linear to $f_{d}$ and experiences a similar change pattern to that described in Fig~\ref{fig: clip_dynamics}(d).
This allows us to estimate the contact establishment and detachment between the object and the clip by detecting the rising and falling pattern of the contact force.

%% file: sections/sec4_estimation.tex
\section{Contact Estimation}~\label{sec: contat_estimation}
In this section, we study the force interaction between the grasped DLO and the clip fixture in the contact and insertion stage. 
Through analyzing the dependency between the applied force and detected force, we define two indicators to estimate the contact establishment and following changes.
To achieve this, we introduce two significant modifications to the clip-fixing skill outlined in Section~\ref{subsec: process}:
\begin{enumerate}
    \item We redefine every single MP to consist solely of a desired feedforward force $\mathbf{f}$, without controlling linear velocity $\mathbf{\dot{x}}^d$.
    Especially in the contact MP, the object is guided by a pushing force to approach the clip after this modification.
    \item In the insertion MP, instead of applying a constant pushing force, $f_{push}$ in the updated clip fixing skill rises gradually from zero
    \begin{equation}
       f_{push}(0) = 0 \ \text{and} \  \frac{\mathrm{d} f_{push}(t)}{\mathrm{d} t} > 0\textbf{.}
    \end{equation}
    This modification extends the interaction period between the DLO and the clip, resulting in a smoother and longer interaction.
\end{enumerate}

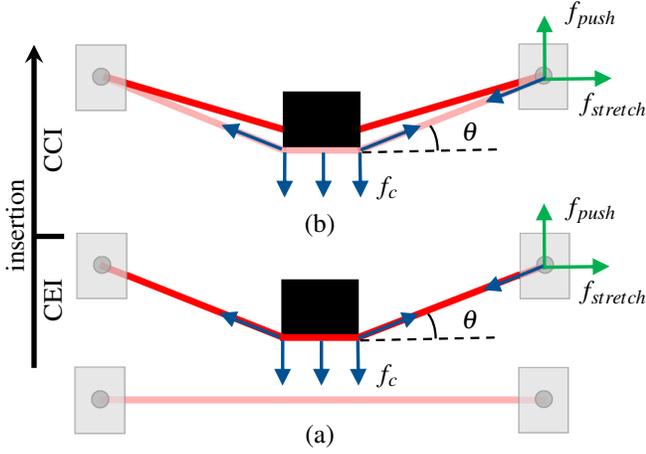
\begin{figure}[t]
    \centering
    \begin{tikzpicture}
    \node[inner sep=0pt] (russell) at (-1.0,0)
    {\includegraphics[width=0.44\textwidth]{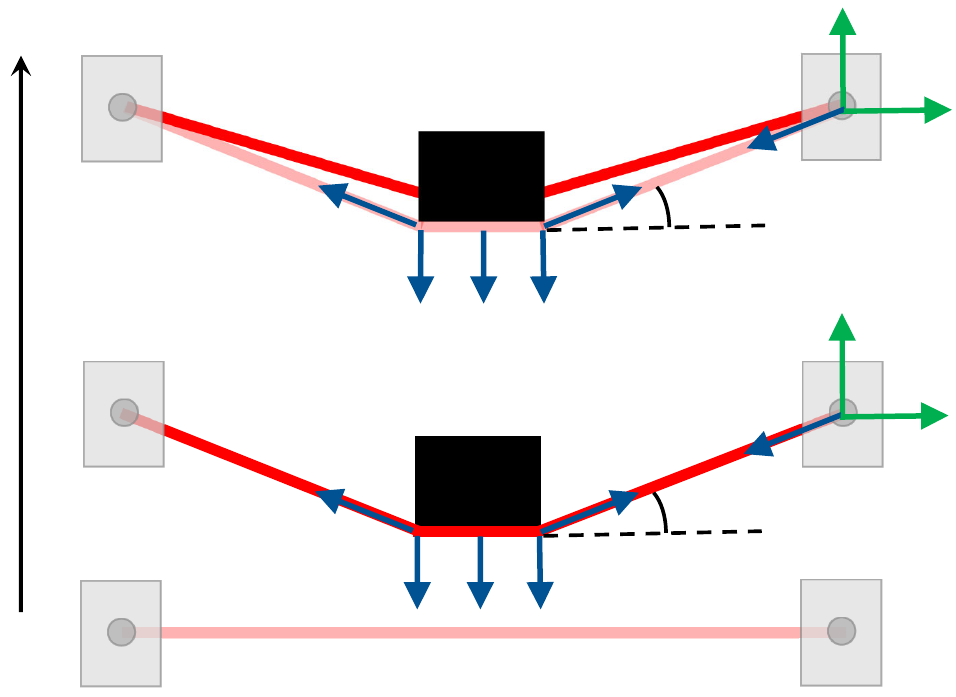}};
    \node at (-1.0, 0) {(b)};
    \node at (2.6,2.8) {$f_{push}$};
    \node at (2.9,1.6) {$f_{stretch}$};
    \node at (-0.1, 0.5) {$f_{c}$};
    \node at (1.0,1.25) {$\theta$};
    
    \node at (-1.0,-2.8) {(a)};
    \node at (2.6,0.2) {$f_{push}$};
    \node at (2.9,-0.9) {$f_{stretch}$};
    \node at (-0.1,-2) {$f_{c}$};
    \node at (1.0,-1.25) {$\theta$};
    
    \draw[draw=white, fill=white] (-4.95,-3) rectangle ++(0.4,6);
    \node[rotate=90] at (-5.05,0) {insertion};
    \draw [-stealth][line width=2pt] (-4.8,-1.9)--(-4.8,2.4);
    \node[rotate=90] at (-4.55,1) {CCI};
    \draw [-][line width=2pt] (-4.8,-0.15)--(-4.4,-0.15);
    \node[rotate=90] at (-4.55,-1) {CEI};
    
    \end{tikzpicture}
    \caption{Top view of cable dynamics in contact with the clip (a) from stretching to contact establishment, and (b) from contact establishment to fixed-in. The clip is represented by the black block in the center. The black arrow on the left describes insertion direction.}
    \label{fig: cable_dynamics_single}
    \vspace{-1.7em} 
\end{figure}

With the modifications above, in all the MPs after stretching, the grasped DLO is stretched by $\mathbf{f_{stretch}}$ and at the same time pushed by $\mathbf{f_{push}}$ into the clip, as shown in Fig~\ref{fig: cable_dynamics_single}. 
Once it establishes contact with the clip, it is under another contact force $\mathbf{f_c}$.
Taking all the external forces applied on the grasped DLO into account, the general dynamics of the DLO can be described as
\begin{equation} \label{equ: general_cable_dynamics_vector}
    m \cdot \mathbf{\ddot{x}}(t) = \mathbf{f_{stretch}} + \mathbf{f_{push}} - \mathbf{f_{c}}(t)\textbf{.}
\end{equation}
Since the DLO is already stretched at both ends to be tension beforehand, the contribution of $\mathbf{f_{stretch}}$ to the acceleration $\mathbf{\ddot{x}}$ can be ignored.
We then consider only the components in $y$ direction in~\eqref{equ: general_cable_dynamics_vector}:
\begin{equation} \label{equ: general_cable_dynamics_scalar}
    m \cdot \ddot{x}(t) = f_{push} -f_{c}(t)\textbf{.}
\end{equation}
In essence,~\eqref{equ: general_cable_dynamics_scalar} can be conceptualized as a system that takes $f_{push}$ as input and generates outputs in the form of $\ddot{x}$ and $f_c$. 
We define a contact establishment indicator (CEI) and a contact change indicator (CCI) to estimate the initial contact establishment and following contact changes respectively by analyzing the interrelationship between the input and output.

\subsection{Contact Change Indicator (CCI)}
As shown in Fig~\ref{fig: cable_dynamics_single}(a), the insertion MP describes the process after the DLO has established a contact with the clip until it is inserted in and the contact terminates.
Inspired by the definition of stiffness (ratio of the resulting deformation to the applied force), we define an indicator for describing the contact change in this process as the rate of change of the resulting contact force to the feedforward force:
\begin{equation}
    \rho_{c} = \frac{\mathrm{d} f_c}{\mathrm{d} f_{push}}\textbf{.}
\end{equation}

To establish a robust relationship between $f_c$ and $f_{push}$, we set two prerequisites for the insertion MP.
Firstly, the DLO is already in a solid contact with the clip, so the deformation of DLO can be neglected, i.e., $\theta$ is almost a constant.
In addition, after the contact MP, robots are forced to pause moving until the velocity is close to zero before the insertion MP starts.
By combining these prerequisites with the second modification we introduced before, we ensure that at each time point between establishing contact and being inserted into the clip, the DLO can be approximated as quasi-static and the acceleration could be neglected so that $\rho_{c} \approx 1$.

As $f_{push}(t)$ rises, an abrupt drop in $\rho_{c}$ occurs at the moment when the DLO is inserted into the clip and the contact disappears that $f_c \approx 0$.
To capture this moment, we make prediction of $\rho_{c}$ in the future:
at each time point $t$, we consider $\rho_{c}(t)$ as a random variable following a Gaussian distribution
\begin{equation}
    \rho_{c}(t) \sim \mathcal{N}(\mu_{t-1},\,\sigma_{t-1}^{2})\textbf{,}
\end{equation}
where $\mu_{t-1}$ and $\sigma_{t-1}$ represent the cumulative average and standard deviation until time step $t-1$, respectively.
When the contact remains stable, $\rho_{c}(t)$ should conform to our prediction.
The instant of a contact change, whether a termination or a new establishment, is detected when $\rho_{c}(t)$ deviates from the prediction, i.e., when it falls outside a confidence interval (CI) specified by the $z$-score.
The condition for contact detachment is formulated as:
\begin{equation}~\label{equ: detach_condition}
    \rho_{c}(t) < \mu_{t-1} - \mathit{Z}\cdot\sigma_{t-1}\textbf{.}
\end{equation}
Similarly, the re-establishment condition is formulated as:
\begin{equation}~\label{equ: reestablish_condition}
    \rho_{c}(t) > \mu_{t-1} + \mathit{Z}\cdot\sigma_{t-1}\textbf{.}
\end{equation}


Capturing contact change using CCI offers a distinct advantage in terms of adaptability when compared to relying on a constant contact force threshold $F_c$.
This adaptability enables CCI to be effectively employed with clips of various size or materials. 
Furthermore, due to its independence from the specific rising pattern of $f_{push}(t)$, $\rho_{c}$ outperforms the contact force change rate $\frac{df_c}{dt}$ which is also differential-based.
These advantages will be evaluated later in Section~\ref{sec: contact_exp}.
\vspace{-0.5em} 
\subsection{Contact Establishment Indicator (CEI)}
Given the first prerequisite of the insertion MP that $\theta$ should be quasi-static, we define an indicator for describing whether there is a solid contact established between the clip and the DLO in the contact MP.
This contact establishment indicator, denoted as $\rho_E$, is defined as the ratio of the contact force to the feedforward force:
\begin{equation}
    \rho_{e} = \frac{f_c}{f_{push}}\textbf{.}
\end{equation}
In theory, the moment when contact is established can be detected by simply measuring whether $f_c>0$.
In practice, however, the contact force detected by robots $f^{ext}_c(t)$ is usually non-zero as it includes additionally some noise and especially measurement error arising from acceleration $ f^{ext}_c(t) =  f_c(t) + m_e\cdot \ddot{x}(t)$, as is shown in Fig~\ref{fig: cable_dynamics_single}(b).
Before any contact is established, the acceleration $ \ddot{x}(t) = \frac{f_{push}}{m}$ is relatively high and thus the measurement error cannot be ignored.
This leads to the modified form of~\eqref{equ: general_cable_dynamics_scalar}:
\begin{equation}\label{equ: no_contact_dynamics}
    (m + m_e)\cdot \ddot{x}(t) = f_{push} - f_c(t)\textbf{.}
\end{equation}
We assume that $f_{push}$ remains constant during this stage and is smaller than the maximum contact force provided by the clip deformation:
\begin{equation} \label{equ: contac_force_constraint}
    \frac{\mathrm{d} f_{push}}{\mathrm{d} t} = 0 \ \text{and}\ f_{push} < \max f_c(t)\textbf{.}
\end{equation}
As the deformation of the clip grows, both $f_c(t)$ and CEI rises.
The second-order differential system in~\eqref{equ: no_contact_dynamics} will eventually reach an equilibrium point where $f_c(t) = f_{push}$ and $\rho_{e}=1$.
This marks the moment when $\theta$ becomes stable and can be considered as the establishment of a solid contact.
In practice, we formulate the contact establishment condition with a threshold $E$ that $\rho_E > E$.


%% file: sections/sec5_enhance.tex
\section{Enhanced Shape Control}
\begin{algorithm}[t!]
\caption{EnhancedShapeControl ($\Psi = \{\mathbf{\psi_i}\}$)}\label{alg: shape_control}
\small{
\begin{algorithmic}[1]
\State \textbf{Initialize} $(S_t, \mathbf{x}_h, f_{ext})$
\For{$\mathbf{\psi_i} \in \Psi$}  
    \If{$\mathbf{x}_h \approx\mathbf{\psi_i}$} \Comment{Start Clip Fixing Iteration}
        \While{$\mathbf{\zeta_t} \neq \mathbf{\zeta^*}$} 
        \State \textbf{Initialize} $\mathbf{\zeta}_t = [0]$
        \State \textbf{Sample} $x_h^z$ and $F_{push}$
        \While {\textbf{not $ExitCondition$}}
        \State $\mathbf{f^{ext}_c} = MP(\mathbf{x_h}, \mathbf{f_{push}})$
        \State $\rho_E, \rho_T = ContactIndicator(\mathbf{f^{ext}_c}, \mathbf{f_{push}}, E, Z)$
        \State $\mathbf{\zeta_t} \leftarrow ContactStateTransiton(\rho_E, \rho_T)$
        \If {$\mathbf{\zeta_t} = \mathbf{\zeta^*}$} 
            \State\textbf{break}
        \EndIf
        \EndWhile
        \State $x_h^z, F_{push} = SampleParam( \mathbf{\zeta^*} - \mathbf{\zeta_t})$
        \EndWhile
        
        \Else \Comment{Start Shape Tracking Skill}
        \State $\mathbf{x}_h \leftarrow ShapeTracking(S_t, \mathbf{\psi_i}, \mathbf{x}_h)$ 
    \EndIf
\EndFor
\end{algorithmic}}
\end{algorithm}
Based on CEI and CCI, the ideal clip fixing process as well as anomalies introduced in subsection~\ref{subsec: process} can be characterized by the contact force pattern, more specifically, as sequences of contact establishment and detachment, denoted as $\mathbf{\zeta}$.
We define the initial contact state after stretching as $0$.
For an ideal clip fixing process, upon the first contact establishment detected by CEI, the contact state turns to $1$.
After that, once the contact detachment is detected by CCI using~\eqref{equ: detach_condition}, the contact state turns to $0$ and the process terminates with the resulting contact state sequence $\mathbf{\zeta}^*=[0, 1, 0]$.
Otherwise, the skill is forced to exit if it reaches a time or displacement limit.
Especially in an overforce movement case, the re-establishment of contact with the rear part of the clip is detected by CCI using~\eqref{equ: reestablish_condition} and the contact state turns again to $1$. 
The contact state sequences of each MP and anomaly is depicted in Fig~\ref{fig: overview}.

We then combine the improved clip fixing skill with the shape tracking skill developed in our prior work~\cite{chen2023contactaware} to form an enhanced adaptive shape control framework which could detect and correct anomalies automatically based on feedback provided by contact sequences.
The two robots collaborate in a ``master-slave" mode.
The master holds one end of the DLO for the whole process, the pose of which is denoted as $\mathbf{x_h}$, and the follower only comes to grasp the DLO when clip-fixing skill starts.
The shape of DLO $S_t$ is obtained from visual perception. 
The shape control skill plans motion of both robots to each fixture while avoiding collision between robots and with fixtures $\Psi=\{\psi_i\}$.
Arriving at one fixture $\psi_i$, the clip fixing skill starts and runs in iterations.
In the first iteration, MP parameters are sampled randomly from respective uniform distributions.
For simplicity, we parameterize $\mathbf{x_h}$ with its $z$ component $x_h^z$, and $f_{push}(t)$ with the maximum force $F_{push}$ it could reach within a certain period $T_{push}$.
After each iteration, in case when anomalies are detected, the upper and lower ranges of the parameter distributions are updated respectively based on the anomaly type, i.e., based on the difference between $\mathbf{\zeta_t}$ and $\mathbf{\zeta}^*$, and MP parameters ($x_h^z$ and $F_{push}$) are sampled again from the updated distribution.
This process is repeated until $\mathbf{\zeta^*}$ is detected.
Afterwards, the robots move to the next fixture.
The shape control framework enhanced with real-time contact estimation is summarized in Algorithm~\ref{alg: shape_control}.

%% file: sections/sec6_experiment.tex
\section{Experiments}
In this section, we present real-world experiments to evaluate the accuracy of the proposed contact estimation approach and the improvements it brings to the DLO shape control process.
As shown in Fig. \ref{fig: clips}, we use two 7 DOF Franka Emika Panda robots for the experiments, both of which are equipped with joint torque sensors and provide 6-axis force torque estimation at the end-effectors.
Throughout this process, all the fixtures remain anchored to the desk, maintaining a constant pose.

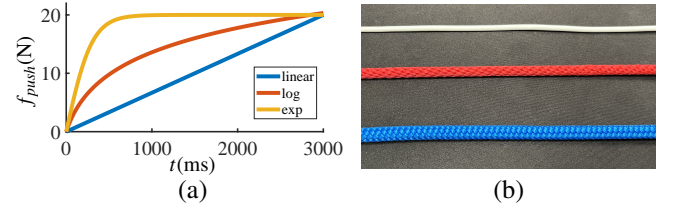
\begin{figure}[t]
    \centering
    \begin{tikzpicture}
    \node[inner sep=0pt] (russell) at (-2.1,0.1)
    {\includegraphics[width=0.23\textwidth]{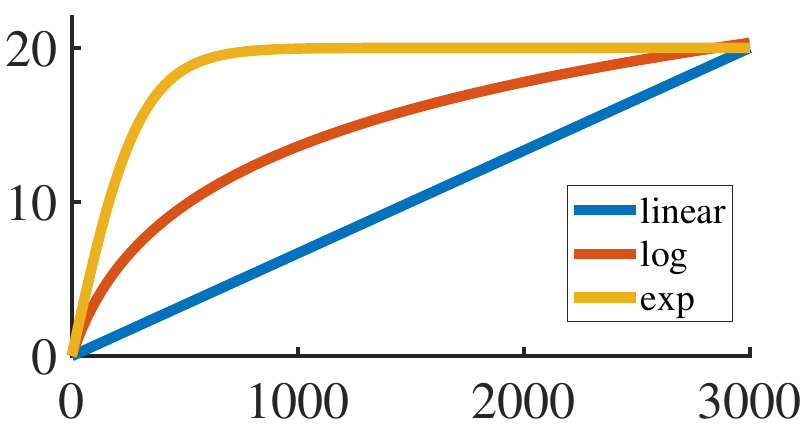}};
    \node at (-2.1,-1.4) {(a)};
    \node at ((-2.1,-1.05) {\footnotesize{$t$(ms)}};
    \node [rotate=90] at ((-4.3,0.2) {\footnotesize{$f_{push}$(N)}};
    \node[inner sep=0pt] (russell) at (2.1,0.0)
    {\includegraphics[width=0.22\textwidth]{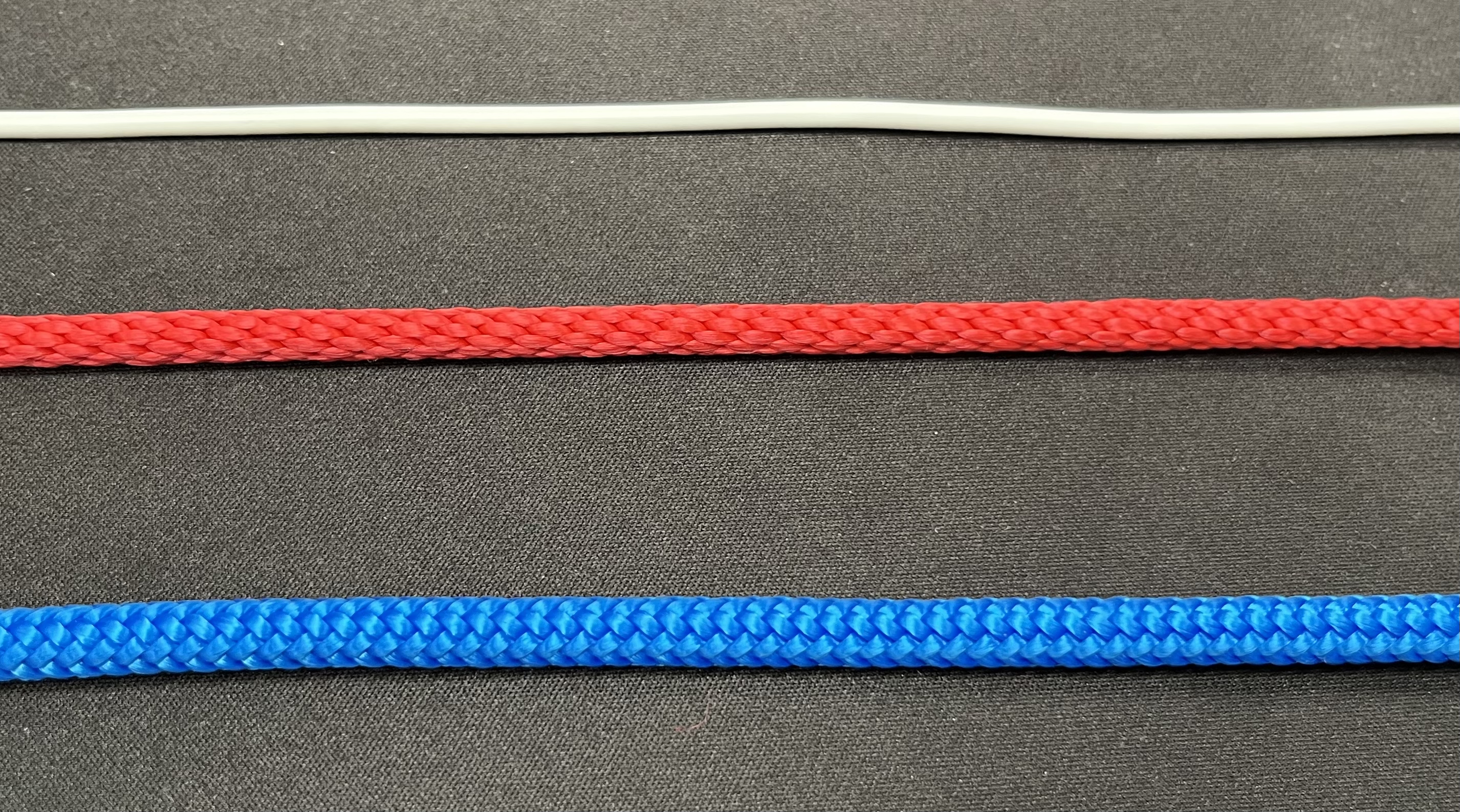}};
    \node at (2.1,-1.4) {(b)};
    \end{tikzpicture}
    \vspace{-1.5em} 
    \caption{Various settings for comparing contact change detection. (a) Different growing patterns of the $f_{push}(t)$. (b) Cables with different radius.}
    \label{fig: ff_rising}
    \vspace{-1.7em} 
\end{figure}



\subsection{Evaluation of Contact Change Detection}\label{sec: contact_exp}
We evaluate the performance of CCI in comparison to two other commonly used indicators for contact change detection, namely, a constant contact threshold $F_c$ and the contact force change rate $\frac{df_c}{dt}$, across various setups.
The detected contact force is smoothened using a Bartlett window of length $50$ ms in real time.
We choose the $99.5\%$ confidence interval with $\mathit{Z}=2.807$ for both $\frac{df_c}{dt}$ and $\rho_{c}$.
In the insertion MP, $\mathbf{f_{push}}$ is supposed to be applied for a duration of $T_{push}=3000$ ms until $F_{push}=20$ N.

\begin{figure*}[t!]
    \centering
    \begin{tikzpicture}
    \node[inner sep=0pt] (russell) at (-4.4,0)
    {\includegraphics[width=0.46\textwidth]{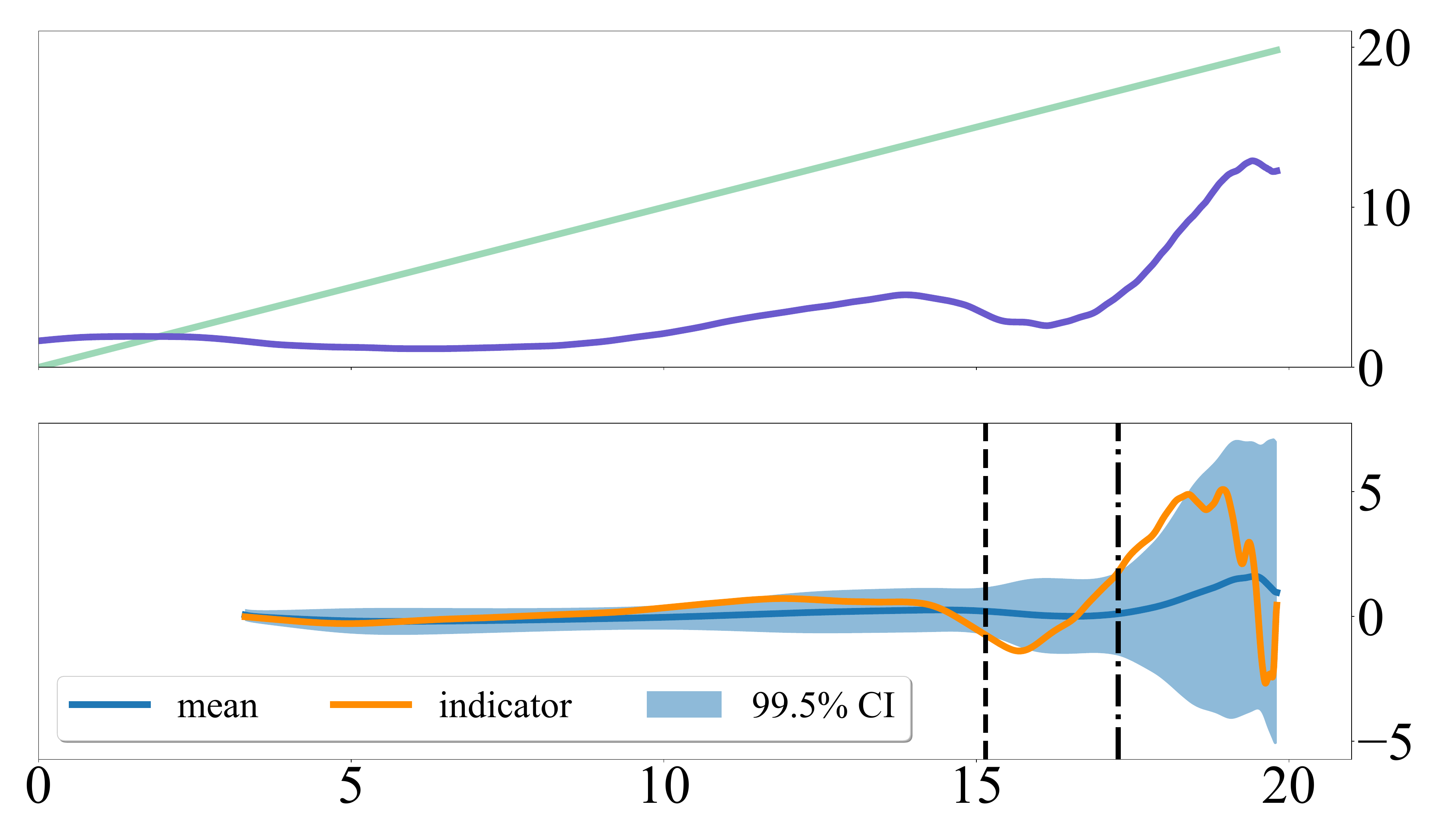}};
    \node[inner sep=0pt] (russell) at (4.4,0)
    {\includegraphics[width=0.46\textwidth]{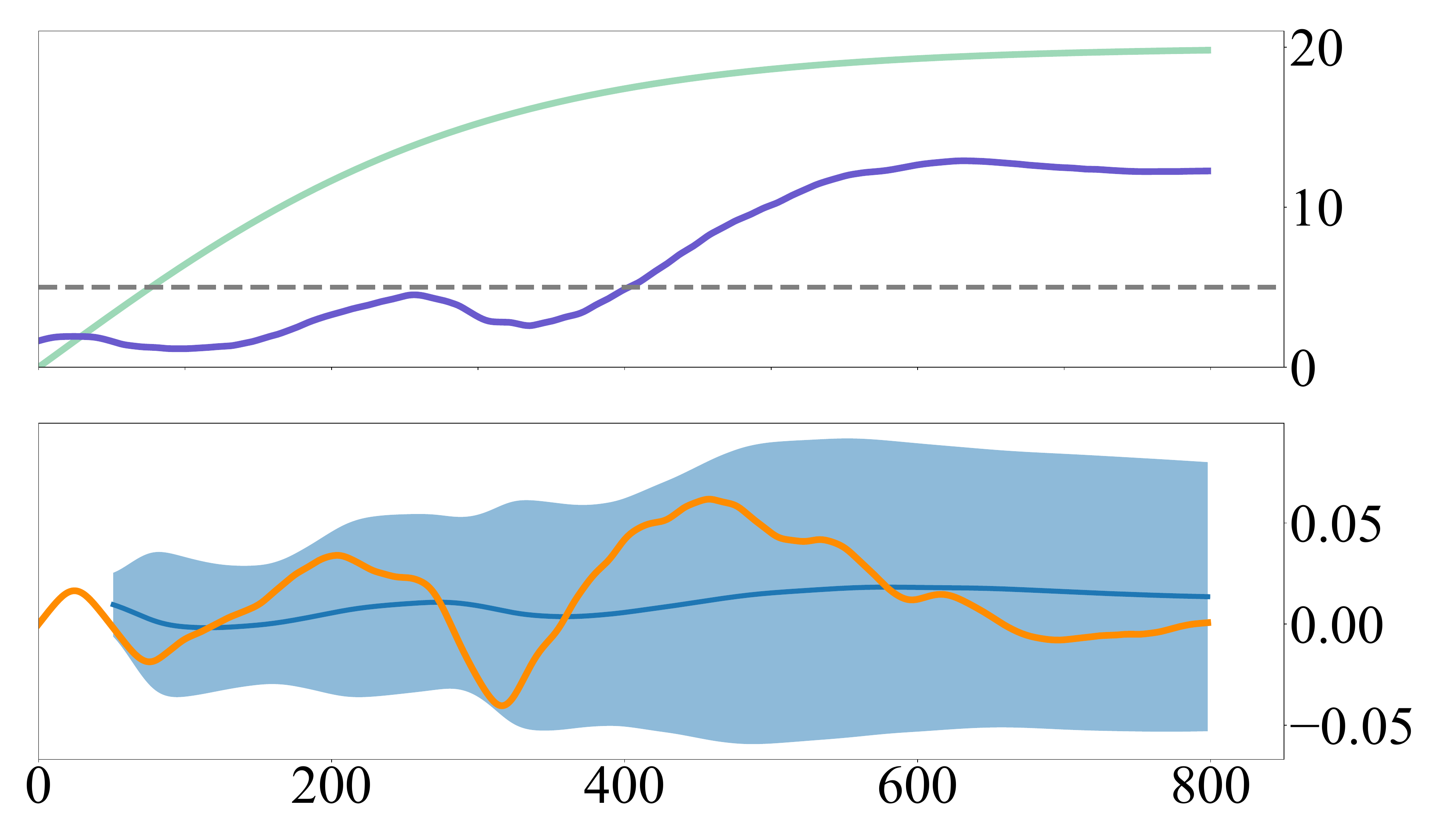}};
    \node at (-4.6,-2.4) {$f_{push}$ (N)};
    \node [rotate=90] at (-8.6,1.3) {$f^{ext}_{c}$ (N)};
    \node [rotate=90] at (-8.6,-0.9) {$\rho_{c}$};
    \node at (-8.6,-1.75) {(a)};
    
    \node at (4.6,-2.4) {$t$ (ms)};
    \node [rotate=90] at (0.2,1.3) {$f^{ext}_{c}$ (N)};
    \node [rotate=90] at (0.2,-0.9) {$\frac{df_c}{dt}$};
    \node at (0.2,-1.75) {(b)};
    \node[color=darkgray] at (7.8, 0.7) {$F_c$};
    \end{tikzpicture}
    \vspace{-0.5em} 
    \caption{Contact estimation under exponentially growing feedforward force. Only the first 800 ms is plotted, as $f_{push}$ remains quasi-constant afterwards. The first row depicts the change of contact force (in purple) to (a) $f_{push}(t)$ and (b) time. The rising of $f_{push}(t)$ is also plotted in green for reference. The second row depicts the change of (a) our contact estimation indicator (CCI) and (b) contact force change rate $\frac{df_c}{dt}$.}
    \label{fig: exp_compare}
    \vspace{-1.7em}
\end{figure*}
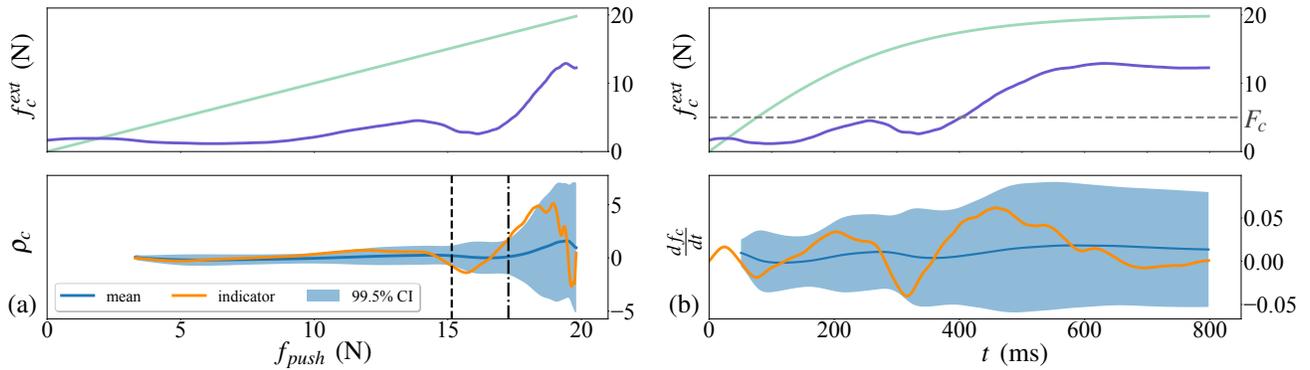

\textbf{Across different rising}
Firstly, we compare performances under three different growing patterns of $f_{push}(t)$, each approximating a linear function, a logarithm function and a exponential function, as shown in Fig~\ref{fig: ff_rising}(a).

We collect contact data by performing overforce movement with every growing pattern in Fig~\ref{fig: ff_rising}(a) at three different fixture poses depicted in Fig~\ref{fig: clips}, namely, P1, P2, and P3, with each setting repeated for 10 times.
The number of successful contact change detection using each indicator is summarized in Table~\ref{tab: change_accuracy_rising}.
Overall, $\rho_{c}$ achieves equal or higher success rate across different rising patterns.

\begin{table}[ht]
\caption{Contact change detection accuracy I.}
\centering
\resizebox{0.48\textwidth}{!}{
\begin{tabular}{p{1cm}|c|c|c|c|c|c|c|c|c}
 \hline
 & \multicolumn{3}{|c|}{$F_c$} & \multicolumn{3}{|c|}{$df_c/dt$} & \multicolumn{3}{|c}{$\rho_{c}$ (\textbf{ours})}\\
 \hline
        &P1 &P2 &P3 &P1 &P2 &P3 &P1 &P2 &P3\\
 \hline
 linear &10 &10 &9  &10 &10 &9  &10 &10 &9 \\
 log    &10 &10 &9  &10 &10 &5  &10 &10 &9 \\
 exp    &10 &1  &4  &9  &1  &2  &10 &10 &5 \\
 \hline
 success &1.0 &0.7 &0.73  &0.96 &0.7  &0.53  &\textbf{1.0} &\textbf{1.0} &\textbf{0.83} \\
 \hline
\end{tabular}}\label{tab: change_accuracy_rising}
\vspace{-1em}
\end{table}

We also look into one trial with exponential rising where $\frac{df_c}{dt}$ fails to detect the contact change instant.
As shown in Fig~\ref{fig: exp_compare}(a), $\frac{df_c}{dt}$ in this case becomes very fluctuating and is unable to reveal the contact change clearly.
In contrast, CCI is able to capture the contact termination accurately at $297$ ms (dashed vertical line) as well as the new establishment at $392$ ms (dashdotted vertical line), as is shown in Fig~\ref{fig: exp_compare}(b).  
The moments when $\rho_{c}$ exceeds the confidence interval are clearly observable.

\textbf{Across different cables and clips}
Furthermore, we compare their performance on cables with different radius (Fig~\ref{fig: ff_rising}(b)) and clips of different sizes and opening directions (Fig~\ref{fig: clips}). 
The number of successful contact change detection using each indicator is listed in Table~\ref{tab: change_accuracy_setting}.
We can observe that when the DLO radius is much larger than the opening of the clip, all three indices achieve high success rate.
As the radius becomes smaller and the contact change turns less obvious, e.g. in the case of cable S with clip C1, $\rho_{c}$ preserves the most robust performance.
In an extreme case where the cable radius is almost the same as the clip opening (S with U1), the contact force is too low to be detected by any indicator.

\begin{table}[ht]
\caption{Contact change detection accuracy II.}
\centering
\resizebox{0.42\textwidth}{!}{
\begin{tabular}{p{0.9cm}|c|c|c|c|c|c|c|c|c}
 \hline
 & \multicolumn{3}{|c}{$F_c$} & \multicolumn{3}{|c}{$df_c/dt$} & \multicolumn{3}{|c}{$\rho_{c}$ (\textbf{ours})}\\
 \hline
    &C1 &C2 &U1 &C1 &C2 &U1 &C1 &C2 &U1 \\
 \hline
 L  &10   &0   &10 &10 &10 &10 &10 &10 &10 \\
 M  &10  &0  &10 &10 &10 &10 &10 &10 &10 \\
 S  &4   &0   &0 &4 &0 &0  &10 &10  &0 \\
 \hline
 success & \multicolumn{3}{|c}{0.48} & \multicolumn{3}{|c}{0.71} & \multicolumn{3}{|c}{\textbf{0.88}} \\
 \hline
\end{tabular}} \label{tab: change_accuracy_setting}
\end{table}


\vspace{-0.5em}
\subsection{Evaluation of Enhanced Shape Control}
\begin{figure}[t]
    \centering
    \begin{tikzpicture}

     \node[inner sep=0pt] (russell) at (0.25,0)
    {\includegraphics[width=0.447\textwidth]{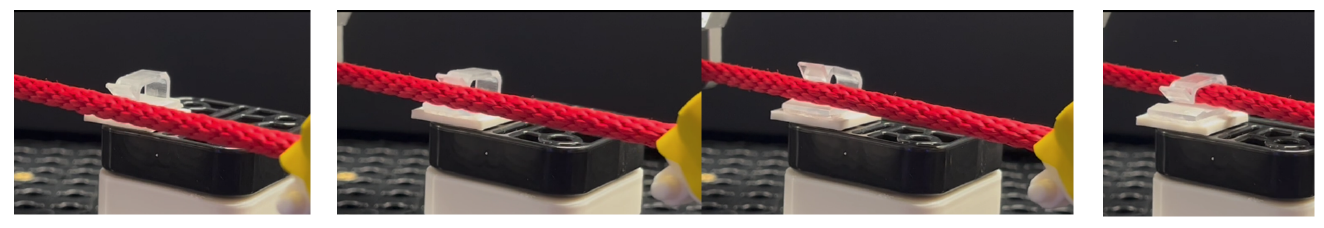}};
    
    \node[inner sep=0pt] (russell) at (0.4,-1.8)
    {\includegraphics[width=0.46\textwidth]{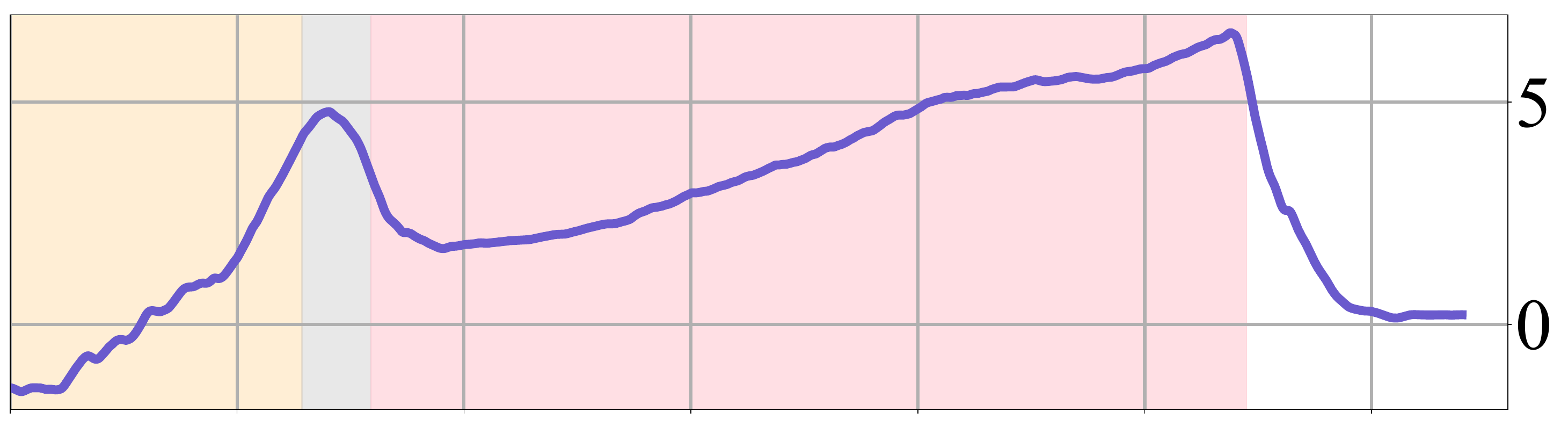}};
    \node [rotate=90] at (-3.9,-1.8) {$f^{ext}_{c}$(N)};

    \node at (-2.8, -3) {contact};
    \node at (0.6, -3) {insertion};
    \node at (3.5, -3) {fixed};
    
    \end{tikzpicture}
    \caption{Process and contact force in a trial of ideal clip fixing. The DLO firstly establishes contact with the clip (yellow), pauses until the velocity decreases (gray), then gets inserted into the clip (pink) and finally remains fixed there (white). }
    \label{fig: trial}
    \vspace{-1.7em} 
\end{figure}
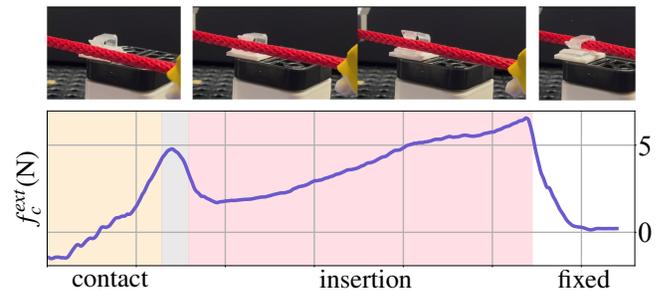

Finally, we evaluate the improvements our contact estimation approach brings to the shape control framework by comparing the success rate with and without contact estimation integrated.
As is shown in Fig~\ref{fig: clips}, four fixtures of three different types are mounted securely on the harness board. 
Each fixture is designed to have a slightly different offset $\delta_z$ in its $z$ axis.
These offsets are hard to be detected by visual observations but may lead to an anomaly in clip fixing.
The framework is aware of the clips' opening direction and originally assumes that all the fixtures are lying on a plane where all $\delta_z = 0$.
For contact MP, we apply $f_{push}=6$ and select $E=0.75$.
The results of clip fixing at each fixture are listed in Table~\ref{tab: shape_control}.
For the complete shape control process, please refer to the accompanying video.
\begin{table}[ht]
\caption{Shape control experiments}
\centering
\resizebox{0.48\textwidth}{!}{
\begin{tabular}{p{1.1cm}|c|c|c|c}
 \toprule
    &U1 ($\delta_z=$-10mm) &C1 ($\delta_z=$3mm) &C3 ($\delta_z=$5mm) &C3 ($\delta_z=$0mm)\\
 \hline
 \textbf{With}  &\textbf{Success}   &\textbf{Success}   &\textbf{Success} & \textbf{Success}\\
 Without  &Missed Contact & Entry Blockage &Entry Blockage & Success\\
 \bottomrule
\end{tabular}} \label{tab: shape_control}
\vspace{-1.7em} 
\end{table}


    
    
    

%% file: sections/sec7_conclusion.tex
\section{Conclusion}
We introduce a contact state estimation approach for shape control task of DLOs using small environmental fixtures. 
Our method, based on two force-derived indices, one for detecting contact establishment and the other for identifying abrupt contact changes, is computationally efficient and ensures real-time performance. 
With this contact estimation approach integrated into the control loop, the shape control framework is able to detect and correct various failures.
Real-world experiments demonstrate the robustness of our approach across diverse experimental setups.
It also enhances the success rate of DLO shape control tasks significantly.


%% file: main.bbl
\begin{thebibliography}{10}
\providecommand{\url}[1]{#1}
\csname url@samestyle\endcsname
\providecommand{\newblock}{\relax}
\providecommand{\bibinfo}[2]{#2}
\providecommand{\BIBentrySTDinterwordspacing}{\spaceskip=0pt\relax}
\providecommand{\BIBentryALTinterwordstretchfactor}{4}
\providecommand{\BIBentryALTinterwordspacing}{\spaceskip=\fontdimen2\font plus
\BIBentryALTinterwordstretchfactor\fontdimen3\font minus
  \fontdimen4\font\relax}
\providecommand{\BIBforeignlanguage}[2]{{%
\expandafter\ifx\csname l@#1\endcsname\relax
\typeout{** WARNING: IEEEtran.bst: No hyphenation pattern has been}%
\typeout{** loaded for the language `#1'. Using the pattern for}%
\typeout{** the default language instead.}%
\else
\language=\csname l@#1\endcsname
\fi
#2}}
\providecommand{\BIBdecl}{\relax}
\BIBdecl

\bibitem{chen2023contactaware}
K.~Chen, Z.~Bing, F.~Wu, Y.~Meng, A.~Kraft, S.~Haddadin, and A.~Knoll,
  ``Contact-aware shaping and maintenance of deformable linear objects with
  fixtures,'' in \emph{2023 IEEE/RSJ International Conference on Intelligent
  Robots and Systems (IROS)}, 2023, pp. 1--8.

\bibitem{navas2022wire}
G.~E. Navas-Reascos, D.~Romero, J.~Stahre, and A.~Caballero-Ruiz, ``Wire
  harness assembly process supported by collaborative robots: Literature review
  and call for r\&d,'' \emph{Robotics}, vol.~11, no.~3, p.~65, 2022.

\bibitem{haouchine2018vision}
N.~Haouchine, W.~Kuang, S.~Cotin, and M.~Yip, ``Vision-based force feedback
  estimation for robot-assisted surgery using instrument-constrained
  biomechanical three-dimensional maps,'' \emph{IEEE Robotics and Automation
  Letters}, vol.~3, no.~3, pp. 2160--2165, 2018.

\bibitem{huang2023learning}
Y.~Huang, C.~Xia, X.~Wang, and B.~Liang, ``Learning graph dynamics with
  external contact for deformable linear objects shape control,'' \emph{IEEE
  Robotics and Automation Letters}, 2023.

\bibitem{zhu2019robotic}
J.~Zhu, B.~Navarro, R.~Passama, P.~Fraisse, A.~Crosnier, and A.~Cherubini,
  ``Robotic manipulation planning for shaping deformable linear objects with
  environmental contacts,'' \emph{IEEE Robotics and Automation Letters},
  vol.~5, no.~1, pp. 16--23, 2019.

\bibitem{huo2022keypoint}
S.~Huo, A.~Duan, C.~Li, P.~Zhou, W.~Ma, H.~Wang, and D.~Navarro-Alarcon,
  ``Keypoint-based planar bimanual shaping of deformable linear objects under
  environmental constraints with hierarchical action framework,'' \emph{IEEE
  Robotics and Automation Letters}, vol.~7, no.~2, pp. 5222--5229, 2022.

\bibitem{jin2022robotic}
S.~Jin, W.~Lian, C.~Wang, M.~Tomizuka, and S.~Schaal, ``Robotic cable routing
  with spatial representation,'' \emph{IEEE Robotics and Automation Letters},
  vol.~7, no.~2, pp. 5687--5694, 2022.

\bibitem{waltersson2022planning}
G.~A. Waltersson, R.~Laezza, and Y.~Karayiannidis, ``Planning and control for
  cable-routing with dual-arm robot,'' in \emph{2022 International Conference
  on Robotics and Automation (ICRA)}.\hskip 1em plus 0.5em minus 0.4em\relax
  IEEE, 2022, pp. 1046--1052.

\bibitem{suberkrub2022feel}
F.~S{\"u}berkr{\"u}b, R.~Laezza, and Y.~Karayiannidis, ``Feel the tension:
  Manipulation of deformable linear objects in environments with fixtures using
  force information,'' in \emph{2022 IEEE/RSJ International Conference on
  Intelligent Robots and Systems (IROS)}.\hskip 1em plus 0.5em minus
  0.4em\relax IEEE, 2022, pp. 11\,216--11\,222.

\bibitem{jansen2009surgical}
R.~Jansen, K.~Hauser, N.~Chentanez, F.~Van Der~Stappen, and K.~Goldberg,
  ``Surgical retraction of non-uniform deformable layers of tissue: 2d robot
  grasping and path planning,'' in \emph{2009 IEEE/RSJ International Conference
  on Intelligent Robots and Systems}.\hskip 1em plus 0.5em minus 0.4em\relax
  IEEE, 2009, pp. 4092--4097.

\bibitem{yao2023estimating}
S.~Yao and K.~Hauser, ``Estimating tactile models of heterogeneous deformable
  objects in real time,'' in \emph{2023 IEEE International Conference on
  Robotics and Automation (ICRA)}.\hskip 1em plus 0.5em minus 0.4em\relax IEEE,
  2023, pp. 12\,583--12\,589.

\bibitem{erickson2017does}
Z.~Erickson, A.~Clegg, W.~Yu, G.~Turk, C.~K. Liu, and C.~C. Kemp, ``What does
  the person feel? learning to infer applied forces during robot-assisted
  dressing,'' in \emph{2017 IEEE International Conference on Robotics and
  Automation (ICRA)}.\hskip 1em plus 0.5em minus 0.4em\relax IEEE, 2017, pp.
  6058--6065.

\bibitem{wang2022visual}
Y.~Wang, D.~Held, and Z.~Erickson, ``Visual haptic reasoning: Estimating
  contact forces by observing deformable object interactions,'' \emph{IEEE
  Robotics and Automation Letters}, vol.~7, no.~4, pp. 11\,426--11\,433, 2022.

\bibitem{wi2022virdo}
Y.~Wi, P.~Florence, A.~Zeng, and N.~Fazeli, ``Virdo: Visio-tactile implicit
  representations of deformable objects,'' in \emph{2022 International
  Conference on Robotics and Automation (ICRA)}.\hskip 1em plus 0.5em minus
  0.4em\relax IEEE, 2022, pp. 3583--3590.

\bibitem{johannsmeier2019framework}
L.~Johannsmeier, M.~Gerchow, and S.~Haddadin, ``A framework for robot
  manipulation: Skill formalism, meta learning and adaptive control,'' in
  \emph{2019 International Conference on Robotics and Automation (ICRA)}.\hskip
  1em plus 0.5em minus 0.4em\relax IEEE, 2019, pp. 5844--5850.

\end{thebibliography}
